\documentclass[manuscript]{acmart}
\AtBeginDocument{%
  \providecommand\BibTeX{{%
    \normalfont B\kern-0.5em{\scshape i\kern-0.25em b}\kern-0.8em\TeX}}}

\usepackage{amsmath}
\usepackage{enumitem}
\usepackage{subcaption}
\usepackage{multirow}
\usepackage{array}
\usepackage{dsfont}
\usepackage{mathtools}
\newcolumntype{P}[1]{>{\centering\arraybackslash}p{#1}}
\usepackage{amsmath, bm}
\usepackage{enumitem}
\usepackage{booktabs}
\usepackage{array, makecell}
\usepackage{booktabs}
\usepackage{placeins}
\usepackage{tabularray}
\usepackage{hhline}
\usepackage{xcolor}
\usepackage{booktabs}
\usepackage{color}
\usepackage{soul}

\usepackage{multirow}
\usepackage[normalem]{ulem}
\useunder{\uline}{\ul}{}

\newlength{\ColWidthNormal} 
\newlength{\ColWidthRowHeader} 
\newlength{\RuleOffsetLeft} 
\newlength{\RuleThicknessNormal} 
\newcolumntype{C}{>{\centering\arraybackslash\leavevmode}p{\ColWidthNormal}}

\setcopyright{acmlicensed}
\copyrightyear{2018}
\acmYear{2018}
\acmDOI{XXXXXXX.XXXXXXX}

\acmConference[Conference acronym 'XX]{Make sure to enter the correct
  conference title from your rights confirmation emai}{June 03--05,
  2018}{Woodstock, NY}
\acmISBN{978-1-4503-XXXX-X/18/06}

\begin{document}

\title{Towards Effective Time-Aware Language Representation: Exploring Enhanced Temporal Understanding in Language Models}


\author{Jiexin Wang}
\email{jiexinwang@scut.edu.cn}
\orcid{0000-0002-7064-6507}
\affiliation{%
  \institution{South China University of Technology}
  \country{China}
  }

\author{Adam Jatowt}
\orcid{0000-0001-7235-0665}
\affiliation{%
  \institution{University of Innsbruck}
  \country{Austria}
  }
\email{adam.jatowt@uibk.ac.at}

\author{Yi Cai}
\authornote{Corresponding author}
\orcid{0000-0002-1767-789X}
\affiliation{%
  \institution{South China University of Technology}
  \country{China}
  }
\email{ycai@scut.edu.cn}

\renewcommand{\shortauthors}{Jiexin Wang, Adam Jatowt, Yi Cai}

\begin{abstract}

In the evolving field of Natural Language Processing (NLP), understanding the temporal context of text is increasingly critical for applications requiring advanced temporal reasoning. Traditional pre-trained language models like BERT, which rely on synchronic document collections such as BookCorpus and Wikipedia, often fall short in effectively capturing and leveraging temporal information. To address this limitation, we introduce BiTimeBERT 2.0, a novel time-aware language model pre-trained on a temporal news article collection. BiTimeBERT 2.0 incorporates temporal information through three innovative pre-training objectives: Extended Time-Aware Masked Language Modeling (ETAMLM), Document Dating (DD), and Time-Sensitive Entity Replacement (TSER). Each objective is specifically designed to target a distinct dimension of temporal information: ETAMLM enhances the model's understanding of temporal contexts and relations, DD integrates document timestamps as explicit chronological markers, and TSER focuses on the temporal dynamics of "Person" entities. Moreover, our refined corpus preprocessing strategy reduces training time by nearly 53\%, making BiTimeBERT 2.0 significantly more efficient while maintaining high performance. Experimental results show that BiTimeBERT 2.0 achieves substantial improvements across a broad range of time-related tasks and excels on datasets spanning extensive temporal ranges. These findings underscore BiTimeBERT 2.0's potential as a powerful tool for advancing temporal reasoning in NLP.\footnote{The code is available at \url{https://github.com/WangJiexin/BiTimeBERT2_0}.}

\end{abstract}

\ccsdesc[500]{Computing methodologies~Natural language processing}

\keywords{pre-trained language models, temporal information, temporal QA, news archives}

\maketitle

\section{Introduction}

In recent years, the digitization of news archives has significantly improved our access to historical data. These digital archives offer a comprehensive perspective of past events, societal shifts, and cultural evolution, assisting researchers and historians in identifying pivotal moments and trends that have played a major role in shaping today's society \cite{Korkeamaki:2019:IDD:3295750.3298931}. 

The integration of chronological data into Natural Language Processing (NLP) and Information Retrieval (IR) systems has similarly gained increasing importance, giving rise to specialized subfields such as Temporal Information Retrieval \cite{alonso2007value, campos2015survey}. This domain focuses on improving the relevance and contextual understanding of search results by incorporating the temporal dimensions of both queries and documents, enabling systems to better capture and leverage the time-sensitive nature of information. 
Beyond its core role in information retrieval, temporal information has found applications across a wide range of tasks, opening exciting avenues for research and innovation. Notable examples include web content analysis \cite{calzarossa2015modeling, mackay2017social}, social media trends analysis \cite{chandrasekaran2020topics, khan2020predicting}, temporal reasoning \cite{jain2023language, zhou2020temporal}, timeline summarization \cite{steen2019abstractive, campos2021automatic}, event detection and ordering \cite{strotgen2012event, bookdercz}, event time prediction \cite{wang2021event}, question answering \cite{pasca, wang2020answering}, content trustworthiness analysis \cite{trustworthiness} and semantic change detection \cite{rosin2022temporal, rosin2022time}. These applications demonstrate the significant and varied impact of temporal data across various research areas and real-world scenarios.

Pre-trained language models, such as BERT \cite{devlin2018bert}, T5 \cite{raffel2020exploring}, and GPT-3 \cite{brown2020language}, have significantly advanced NLP by enabling deeper, context-aware linguistic representations. These models excel in various NLP tasks, including text classification, sentiment analysis, and machine translation. However, they often struggle to capture essential domain-specific information, primarily due to their reliance on training with general-purpose corpora like English Wikipedia, and on training objectives not tailored to domain-specific characteristics. In scenarios where understanding temporal dynamics is crucial, most existing models face challenges in effectively integrating temporal information, such as temporal expressions or document timestamps. This capability is particularly vital when processing news articles and the aforementioned applications where time is an important factor. Developing language models that are not only context-aware but also adept at comprehending temporal aspects would lead to a more refined and accurate interpretation of temporal information, thereby significantly enhancing their effectiveness in time-related tasks.

In this study, we present BiTimeBERT 2.0, a novel time-aware language model pre-trained on a temporal news collection with three training objectives: \emph{extended time-aware masked language modeling} (ETAMLM), \emph{document dating} (DD), and \emph{time-sensitive entity replacement} (TSER). 
Each objective targets a distinct aspect of temporal understanding. The first dimension, \emph{content time}, relates to the temporal expressions within the text, shedding light on the temporal context and the period of the events discussed. The second dimension, \emph{document timestamp}, serves as a chronological marker, indicating the creation date of each document and establishing a definite temporal reference. These two explicit temporal dimensions have proven valuable across various applications, such as temporal web search \cite{stack2006full, joho2013survey}, summarization \cite{allan2001temporal, barros2019natsum}, and event ordering \cite{honovich2020machine}, all benefiting from a nuanced understanding of temporal aspects. 
The third aspect, \emph{time-sensitive entity}, focuses on dynamic elements like individuals that often imply specific timeframes. For example, mentioning "Sam Altman" generally suggests a period around 2022 or later, corresponding to the rise of ChatGPT. 
Existing studies \cite{TNER, xu2022time, gonzalez2023injecting, agarwal2018dianed} have shown that "Person" entities are inherently more time-sensitive due to their shorter lifespans or activity durations. Specifically, they reveal that integrating temporal information into tasks like Named Entity Recognition (NER) and Named Entity Disambiguation (NED) yields the most significant performance improvements for "Person" entities, underscoring the close relationship between these entities and specific time periods.
In contrast, while entities like "Location" and "Organization" can also carry temporal relevance (e.g., the site of a historical event), they tend to exhibit a more static nature over time. For instance, a location like "New York" retains its relevance across decades or even centuries, regardless of its association with various events, demonstrating less temporal dynamism compared to individuals, political leaders, or cultural icons. Consequently, integrating temporal information into NER and NED results in only modest improvements for these types of entities. This study, therefore, places particular emphasis on "Person" entities due to their pronounced temporal significance.\footnote{For a detailed discussion of the temporal characteristics of named entities, please refer to the final paragraph of Section \ref{section2_1} in Related Work.}

In the experimental evaluation, we conduct extensive experiments to thoroughly assess the effectiveness of the proposed pre-training tasks. The results from these experiments unequivocally demonstrate that BiTimeBERT 2.0, harnessing the three distinct types of temporal information through its pre-training tasks, yields remarkably effective time-aware representations. Furthermore, our findings underscore the substantial potential of BiTimeBERT 2.0 to significantly enhance performance across a wide spectrum of time-related downstream tasks, showcasing its versatility and applicability in various applications. 

This paper represents a substantial extension of our earlier work, BiTimeBERT, presented in \cite{wang2023bitimebert}. First, we refine the Time-Aware Masked Language Modeling (TAMLM) task into Extended Time-Aware Masked Language Modeling (ETAMLM), which additionally incorporates masking of temporal signals that denote temporal relations, such as "before", "after", and "during". This refinement enables the model to achieve a deeper understanding of temporal contexts, surpassing the previous approach that focused solely on masking explicit temporal information. Second, we introduce the time-sensitive entity replacement (TSER) objective, with a particular focus on dynamic "Person" entities, an underexplored yet critical aspect of temporal information. This addition enriches the model's ability to capture the temporal dynamics associated with individuals, enhancing its versatility in handling time-sensitive content. Third, we implement a pre-training corpus processing strategy that excludes sentences lacking temporal information. This approach allows us to pre-train BiTimeBERT 2.0 on a more specialized temporal news collection, effectively reducing training costs from 80 GPU hours to 38 while preserving high-quality results in most cases.
Furthermore, we investigate two alternative objectives that leverage temporal information in different ways: Time-Sensitive Entity Masked Language Modeling (TSEMLM), which emphasizes masking "Person" entities to enhance temporal dynamics comprehension, and Temporal Relation Word Replacement (TRWR), which involves replacing temporal signals to strengthen the model's temporal reasoning capabilities. 
Additionally, we conduct comprehensive experiments to rigorously evaluate BiTimeBERT 2.0's performance across a variety of time-related tasks. Our findings demonstrate the model's effectiveness and robustness in temporal understanding and temporal reasoning.
Together, these contributions distinguish this work from previous research by providing a more comprehensive framework for advancing time-aware language modeling.

To sum up, this work makes the following contributions:  
\begin{enumerate}[leftmargin=1.5em, labelwidth=!]
\item We thoroughly explore the impact of integrating three distinct types of temporal information into pre-trained language models. Our findings reveal that infusing language models with temporal information improves their performance across various downstream time-related tasks, highlighting the importance of temporal context.

\item We introduce BiTimeBERT 2.0, a novel time-aware language model specifically trained through three pre-training objectives. In addition, we implement several strategic enhancements, including the refinement of time-aware masked language modeling, the introduction of time-sensitive entity replacement objectives, and more efficient training with a focused temporal news corpus, all of which contribute to more effective time-aware representations.

\item We perform extensive experiments across a range of time-related tasks, illustrating BiTimeBERT 2.0's superior ability to capture and leverage temporal information compared to existing models. 
Notably, BiTimeBERT 2.0 excels in generalizing to challenging datasets outside its pre-training temporal scope. These results not only validate our model's enhanced capacity for generating time-aware representations but also showcase its practical effectiveness for real-world applications. 

\end{enumerate}

\section{Related Work}

\subsection{Temporal Dimensions of Text}
\label{section2_1}
According to \citet{campos2015survey}, there are two key temporal dimensions of a document or a query: timestamp and content time (or focus time). The document timestamp refers to the time when the document was created, whereas the content time refers to the time mentioned or implicitly referred to in the document's content (e.g., a document published today\footnote{For reference, we assume "2025/01/01" as the current date.} about WWII would have its timestamp as January 1, 2025, and the content time as 1939-1945). Similarly, the query timestamp indicates the time when the query was issued, and its content time relates to the time period associated with the query (e.g., query like "Winter Olympics 1988" issued three months ago would have its timestamp as October 2024,\footnote{Similarly, we assume "2025/01/01" as the reference date for this context, so a query like "Winter Olympics 1988" issued three months ago would have a timestamp of October 2024.} and would refer to the time period when the winter sports took place in Calgary, Alberta in 1988). Additionally, since a document usually contains sentences related to different events that take place at different time points, the document's content time is typically represented by a set of time intervals \cite{focustime}. 
For readers, timestamp information aids in quickly locating the content on a timeline and assessing the document's up-to-dateness. 
On the other hand, content time can help to strengthen our understanding of particular aspects of the document. For example, the development of events and their causal relations can be understood by analyzing the relations between different content temporal information scattered throughout text content. Recently, exploiting these two types of temporal information in documents and queries has gained increasing importance in IR and NLP. Their interplay can be utilized to develop time-specific search and exploration applications \cite{alonso2011temporal, kanhabua2016temporal}, such as temporal web search \cite{stack2006full, joho2013survey}, temporal question answering \cite{wang2020answering, wang2021improving}, search results diversification \cite{berberich2013temporal, styskin2011recency} and clustering \cite{alonso2009clustering, svore2012creating}, summarization \cite{allan2001temporal, barros2019natsum}, historical event ordering \cite{honovich2020machine} and so on.

Beyond the explicit temporal dimensions of timestamps and content time, named entities inherently encode implicit temporal information, providing additional context for understanding time-specific aspects of text. Previous studies have emphasized the strong correlation between named entities and specific time periods, showing that integrating temporal information can significantly enhance tasks like Named Entity Recognition (NER) and Named Entity Disambiguation (NED) \cite{TNER, xu2022time, gonzalez2023injecting, agarwal2018dianed}. 
Among named entities, "Person" entities consistently exhibit the most substantial performance gains. For instance, \citet{TNER} found that "Person" entities achieved the best performance in NER tasks when incorporating temporal information, with an F1 score approximately 20 points higher than that of "Location" or "Organization" entities across various experimental settings.
Similarly, \citet{agarwal2018dianed} reported that while "Location" and "Organization" entities showed moderate gains in NED accuracy with temporal integration  (5.15\% and 6.49\%, respectively), "Person" entities experienced a dramatic improvement, with accuracy increasing from 9.63\% to 39.91\%. This difference can be attributed to the relatively shorter life spans or well-defined activity durations of "Person" entities, such as political tenures or artistic movements, making them inherently more time-sensitive. 
In contrast, entities like "Location" or "Organization" exhibit greater temporal consistency. For example, "Paris", though associated with notable events such as the 1889 World’s Fair or the 2024 Olympics, maintains its characteristics and relevance over centuries. This enduring relevance reduces its association with specific time periods compared to the more dynamic and time-bound nature of "Person" entities. 
Moreover, while "Event" (e.g., WWII) and "Product" (e.g., iPhone 15) entities also carry implicit temporal references, they pose greater challenges in precise identification. For instance, spaCy's widely used NER model,  en\_core\_web\_sm, achieves an F1-score of 0.8798 for "Person" entities but drops significantly to 0.3421 for "Product" entities and 0.4186 for "Event" entities.\footnote{\href{https://github.com/explosion/spacy-models/blob/3d026eec88c53128ed71e10b399d0084361a11a3/meta/en_core_web_sm-3.8.0.json}{\texttt{Model Metadata for en\_core\_web\_sm.}} Even spaCy's advanced en\_core\_web\_trf model, which enhances "Event" and "Product" entity detection, still lags far behind "Person" entities and comes with significantly higher computational costs.} 
Furthermore, these entities are comparatively rarer in text. We conducted a preliminary experiment utilizing the en\_core\_web\_sm NER model on a dataset of 50,000 randomly sampled news articles from the NYT corpus.\footnote{These articles are excluded from pre-training and reserved for evaluating document timestamp estimation, as detailed in Section \ref{section4_1}.} The results revealed that the model identified 241,917 "Person" entities, starkly contrasting with only 3,893 "Event" entities and 4,975 "Product" entities.
Motivated by these findings, our study places a particular emphasis on "Person" entities, recognizing their intrinsic temporal significance.

\subsection{General Pre-trained Language Models}
The field of Natural Language Processing (NLP) has witnessed a transformative evolution with the advent of pre-trained language models. Early models such as Word2Vec \cite{mikolov2013efficient} and GloVe \cite{pennington2014glove} lay the groundwork by capturing word-level representations, but it is the introduction of contextual models such as ELMo \cite{peters2018deep} that first enable a deeper understanding of word usage within specific contexts. The emergence of Transformer-based models, notably BERT \cite{devlin2018bert}, T5 \cite{raffel2020exploring}, and GPT \cite{radford2018improving}, mark a pivotal shift. These models, leveraging the Transformer architecture \cite{vaswani2017attention}, are typically pre-trained on large general-purpose corpora of text and then fine-tuned for specific NLP tasks, which offer advanced context-aware capabilities.
BERT, in particular, utilizes masked language modeling (MLM) and next sentence prediction (NSP) to enhance understanding of text sequences. RoBERTa \cite{liu2019roberta} further refines BERT's approach by optimizing key hyperparameters and removing the NSP task, while ALBERT \cite{lan2019albert} introduces sentence order prediction (SOP), achieving higher performance with fewer parameters. Further innovations include XLM \cite{lample2019cross}, which incorporates translation language modeling (TLM) for cross-lingual training, XLNet \cite{joshi2020spanbert} with its permuted language modeling (PLM), and SpanBERT \cite{joshi2020spanbert}, which extends BERT by masking contiguous spans and employing a span boundary objective. 
Despite these advancements, a notable limitation of these language models is their reliance on general-purpose text corpora, and training objectives not tailored for specific domains, which restricts their effectiveness in domain-specific applications.

\subsection{Language Models for Specific Domains}

To address the limitations of general-purpose language models, there has been a trend towards developing models tailored to specific domains, leading to enhanced performance in domain-specific tasks. A common approach involves adapting pre-trained models by reapplying their pre-training tasks on domain-specific corpora. For instance, SciBERT \cite{beltagy2019scibert} is trained on scientific texts, BioBERT \cite{lee2020biobert} on biomedical documents, ClinicalBERT \cite{huang2019clinicalbert} on clinical records, and LegalBERT \cite{chalkidis2020legal} on legal documents. These domain-adapted models have shown remarkable effectiveness in their respective fields, such as BioBERT's improved accuracy in biomedical entity recognition and biomedical relation extraction. 
Beyond simple domain adaptation, there is a growing interest on continually pre-training existing models for targeted applications using specific training objectives. For example, SentiLARE, proposed by \citet{ke2019sentilare}, builds upon RoBERTa by introducing a label-aware masked language model to improve sentiment analysis. Similarly, WKLM (Weakly Supervised Knowledge-Pretrained Language Model) \cite{xiong2019pretrained} enhances BERT for entity-centric tasks through an entity replacement objective, which improves the model's comprehension of real-world entities. Another example is the work by  \citet{althammer2021linguistically} who propose linguistically informed masking (LIM) as a domain-specific objective for patent-related tasks, applying continual pre-training to models like BERT and SciBERT. These advancements underscore the importance of refining language models to accommodate the unique characteristics and complexities of specialized domains, enabling more precise and effective language representation for targeted applications.

\subsection{Incorporating Temporal Information into Language Models}
\label{section2_4}

The integration of temporal information into language models has become an increasingly prominent area of research, leading to various innovative approaches aimed at enhancing temporal reasoning and understanding \cite{giulianelli2020analysing, dhingra2021time, rosin2022temporal, rosin2022time, cole2023salient}. A notable example is the work of \citet{dhingra2021time}, which modifies the pre-training process by incorporating timestamp information into the masked language modeling (MLM) objective. This approach facilitates the acquisition of temporally-scoped knowledge, showing improved performance in tasks like question answering. Similarly, \citet{cole2023salient} propose the Temporal Span Masking (TSM) task, an adaptation of Salient Span Masking (SSM) \cite{guu2020retrieval}. TSM specifically focuses on masking temporal expressions in sentences and trains the model to reconstruct these expressions. While these approaches predominantly employ Transformer encoder-decoder architectures, other studies have explored the utility of Transformer encoder-only models. These models have proven effective in semantic change detection, a task that requires the model to identify and understand the extent of semantic shifts in words over time. For instance, \citet{giulianelli2020analysing} introduce the first unsupervised method for semantic change detection, leveraging contextualized embeddings from BERT. Further advancements in this domain include the work of \cite{rosin2022temporal} and \cite{rosin2022time}. The former \cite{rosin2022temporal} enhances the self-attention mechanism \cite{vaswani2017attention} by integrating timestamp data into attention score computations, while the latter \cite{rosin2022time} presents TempoBERT, a time-aware variant of BERT. TempoBERT employs a pre-training objective called Time Masking, which preprocesses input texts by appending timestamps and masking them during training. This approach enables TempoBERT to achieve state-of-the-art results in semantic change detection by focusing on temporal shifts in language.

Despite these advancements, existing approaches face several limitations that limit their applicability and effectiveness. First, most time-aware models focus on either content temporal information or document timestamps, overlooking the necessity of integrating both dimensions for a comprehensive temporal understanding. Second, the pre-training of these models often relies on coarse temporal granularities, such as years or decades, which do not provide the precision necessary for finer-granularity tasks like month-level analysis. This limitation is particularly evident in applications requiring detailed time-sensitive insights, for instance, in web content analysis \cite{calzarossa2015modeling, mackay2017social}. Third, the temporal representations generated by these models are not consistently reliable. For example, \citet{rosin2022time} observed that TempoBERT underperformed compared to a simple fine-tuned BERT model in sentence time prediction tasks, highlighting gaps in its temporal reasoning capabilities.
To address these challenges, our previously proposed model, BiTimeBERT \cite{wang2023bitimebert}, introduced two novel pre-training tasks designed to effectively leverage temporal information. The first objective, Time-Aware Masked Language Modeling (TAMLM), focuses on content time by prioritizing the masking of temporal spans rather than randomly sampling tokens. The second objective, Document Dating (DD), incorporates document timestamps into the pre-training process. To the best of our knowledge, BiTimeBERT represents the first time-aware language model to comprehensively explore and integrate both types of temporal information. 

This extended work enhances BiTimeBERT by incorporating temporal signals through the newly introduced ETAMLM objective, refining the model's ability to understand temporal contexts. Additionally, while the original BiTimeBERT primarily focused on two temporal dimensions (content time and document timestamp), the proposed BiTimeBERT 2.0 expands its scope by emphasizing a third temporal aspect: time-sensitive entities, with a particular focus on "Person" entities, during pre-training. Furthermore, a refined corpus processing strategy significantly reduces training time, achieving a more efficient yet robust pre-training process. Extensive experimental results demonstrate that BiTimeBERT 2.0 provides comprehensive temporal representations and achieves strong performance across a variety of time-related tasks, particularly excelling in datasets that fall outside the temporal scope of its pre-training corpus.

\section{Method}

\begin{figure}[htp]
\centering

\begin{subfigure}{0.99\textwidth}
  \centering
  \includegraphics[width = 0.96\textwidth]{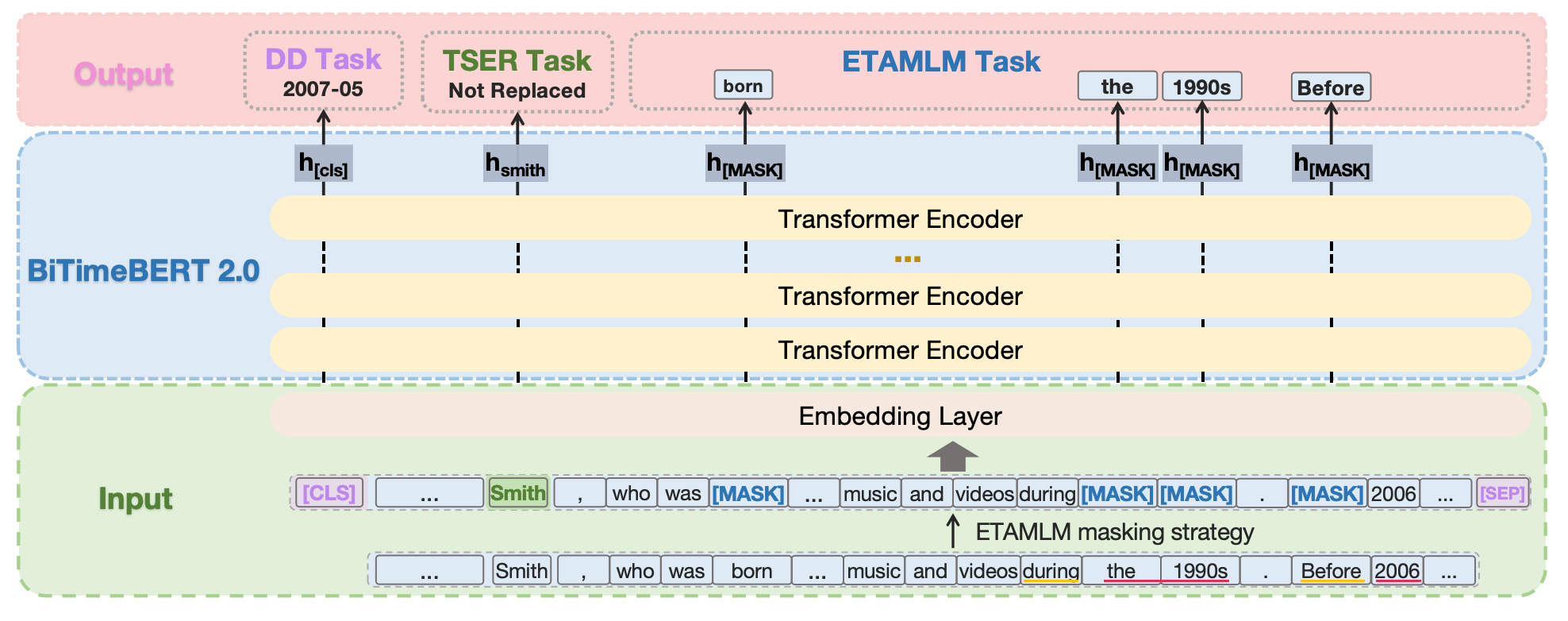}
  \vspace{-.4em}
  \caption{An illustration of BiTimeBERT 2.0 model's pre-training process, which includes the ETAMLM, DD and TSER tasks.}
  \label{fig1}
\end{subfigure}

\vspace{1.5em}

\begin{subfigure}{0.99\textwidth}
  \centering
\centering
  \includegraphics[width = 0.75\textwidth]{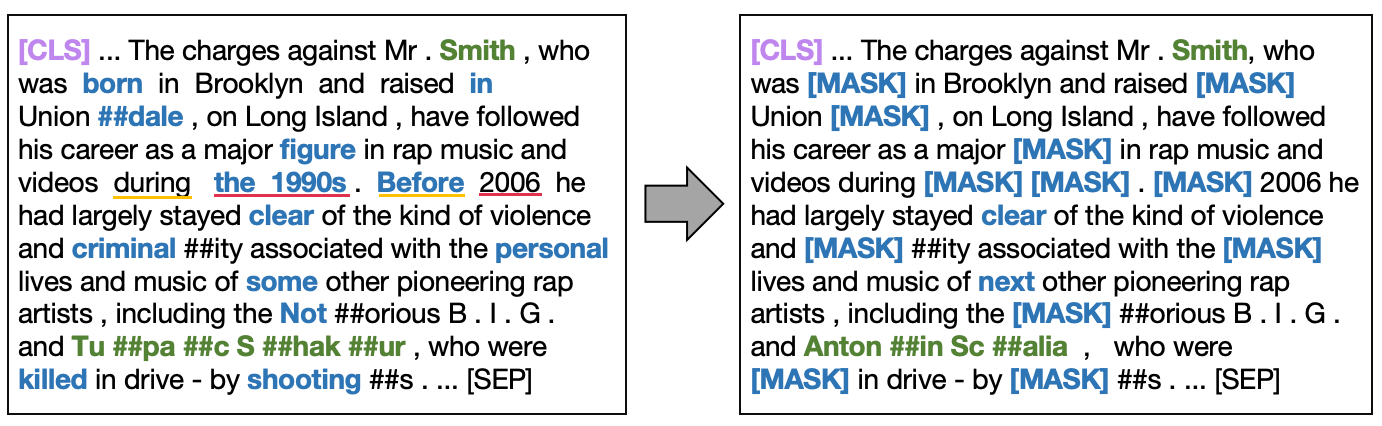}
  \vspace{-.4em}
  \caption{An illustrative input example for the BiTimeBERT 2.0 model during the pre-training phase.}
  \label{fig2}
\end{subfigure}

\caption{A comprehensive overview of the BiTimeBERT 2.0 model's pre-training process, incorporating the ETAMLM, DD, and TSER tasks, along with an illustrative input example during the pre-training phase.}
\label{fig1_fig2}
\end{figure}

In this section, we introduce BiTimeBERT 2.0, the proposed time-aware language model built on the Transformer encoder architecture \cite{vaswani2017attention}, with its training process illustrated in Figure~\ref{fig1}. BiTimeBERT 2.0 differentiates itself from existing language models such as BERT \cite{devlin2018bert} and TempoBERT \cite{rosin2022time} in four critical ways, each contributing to its enhanced time-aware representation capabilities. Firstly, unlike general pre-trained language models that rely on synchronic datasets such as Wikipedia or web crawls, BiTimeBERT 2.0 is trained on a news article collection spanning over two decades.\footnote{The temporal news collection utilized in this study is the New York Times Annotated Corpus (NYT corpus), encompassing over 1.8 million news articles spanning from January 1, 1987, to June 19, 2007.} While models like RoBERTa \cite{liu2019roberta} do include news datasets in their training (e.g., CC-NEWS \cite{nagel2016cc}), they often fail to prioritize or effectively leverage the temporal dimensions of their documents. Furthermore, in comparison with our earlier work on BiTimeBERT \cite{wang2023bitimebert}, which used the entire NYT corpus, BiTimeBERT 2.0 adopts a more focused temporal news collection by deliberately excluding sentences that lack temporal content. This refinement significantly reduces training costs while preserving the high quality of most results.\footnote{Details regarding the pre-training corpus processing are provided in Section \ref{section4_1}.}
Secondly, BiTimeBERT 2.0 employs an innovative \emph{extended time-aware masked language modeling} (ETAMLM) strategy. Unlike traditional random token masking, ETAMLM prioritizes the masking of temporal information within text. This technique not only incorporates the inherent domain knowledge but also enriches the model's comprehension of temporal information embedded in the document content. In contrast to our earlier work, which focused solely on masking temporal expressions, the refined ETAMLM additionally targets temporal signals that indicate relations, such as "before", "after", and "during". This refinement further improves the model's understanding of temporal contexts. 
Third, BiTimeBERT 2.0 replaces the conventional Next Sentence Prediction (NSP) objective with a \emph{document dating} (DD) task, effectively integrating timestamp information during pre-training. As document dating involves time prediction, this objective enriches the model with both time-related and task-oriented knowledge, potentially aiding in improving the performance in other time-sensitive tasks. 
Lastly, this extended work introduces the \emph{time-sensitive entity replacement} (TSER) objective, with a specific emphasis on dynamic "Person" entities. This addition significantly enhances the model's capacity to analyze and interpret temporal dynamics, thereby increasing its overall temporal sensitivity. 
Figure~\ref{fig1} illustrates the three core objectives: ETAMLM, DD, and TSER, whereas Figure~\ref{fig2} provides an illustrative example of the input format during the pre-training phase.\footnote{The example news article, titled "Add a Drunken-Driving Charge to Busta Rhymes's Troubles", was published in The New York Times on May 4, 2007.} BiTimeBERT 2.0 is jointly trained on these tasks, utilizing additional layers integrated with its Transformer network. Each task employs cross-entropy as its loss function.

It is important to note that in our exploration of BiTimeBERT 2.0, we also experiment with two additional pre-training tasks that harness temporal information in different ways, namely \emph{time-sensitive entity masked language modeling} (TSEMLM) and \emph{temporal relation word replacement} (TRWR). TSEMLM is designed to mask "Person" entities in text following the masking of content time. TRWR, on the other hand, involves replacing temporal signals within text: in 50\% of the cases, the temporal signals are substituted with alternatives of a different temporal nature, such as replacing "before" to "after" or "during". Both TSEMLM and TRWR, detailed in Section \ref{section4_4}, are designed to investigate whether incorporating "Person" entities and temporal signals through alternative pre-training objectives could enhance model performance. While these auxiliary tasks demonstrated promising results on certain downstream tasks, our experimental findings reveal that they are less effective compared to BiTimeBERT 2.0. For instance, on the WOTD dataset, BiTimeBERT 2.0 surpasses the model incorporating TSEMLM by 25.58\% in ACC and 19.12\% in MAE and outperforms the model incorporating TRWR by 19.76\% in ACC and 13.83\% in MAE. \footnote{Section \ref{section5_1} provides a more detailed analysis of the experimental results comparing these models.}

\subsection{Extended Time-aware Masked Language Modeling}

The first pre-training objective, Extended Time-Aware Masked Language Modeling (ETAMLM), integrates both content time and temporal signals into the pre-training process. Content time refers to explicit temporal expressions embedded within the document's content, while temporal signals are specific words that indicate temporal relations. Following \cite{jia2018tempquestions}, temporal signals can be categorized into three types based on the relations they indicate: BEFORE, AFTER, and OVERLAP. For instance, BEFORE type includes words like "before" and "prior to", AFTER type includes words such as "after" and "following", and OVERLAP type includes words like "during" and "in". Both content time and temporal signals are essential for comprehending event developments and discerning relationships between events described in text, as evidenced in applications like timeline summarization \cite{yu2021multi} and temporal question answering \cite{wang2020answering, wang2021improving}.

Consider a token sequence $X = (x_{1}, x_{2},..., x_{n})$, where each $x_{i}$ $(1\le i \le n)$ represents a token in the vocabulary. The first step in the ETAMLM process involves identifying temporal expressions within the document content.\footnote{Specifically, we utilize the widely used spaCy NER model, en\_core\_web\_sm, for this identification. During this step, "Person" entities are also identified for use in the latter TSER objective.} 
These temporal expressions, highlighted with red underlines as shown in Figure~\ref{fig1} and Figure~\ref{fig2}, form the set $T = (t_{1}, t_{2},..., t_{m})$, where each $t_{i}$ $(1\le i \le m)$ corresponds to a specific temporal expression found in the text. Next, we locate temporal signals within the text based on a predefined list derived from \cite{jia2018tempquestions} and the extracted temporal expressions. These signals are indicated by orange underlines in Figures~\ref{fig1} and \ref{fig2} and collectively form the temporal signal set $S$. Third, unlike BERT's random masking approach, which uniformly samples 15\% of the sequence tokens, ETAMLM prioritizes masking temporal expressions and temporal signals to enhance temporal comprehension. 
Specifically, we randomly sample 30\% of the entire temporal expressions in $T$ and 30\% of the entire temporal signals in $S$, as exemplified by the masked tokens "the 1990s" and "Before" in Figure~\ref{fig1}. This 30\% masking rate was chosen based on findings from our previous work \cite{wang2023bitimebert}, where it was shown to strike a balance between effectively learning temporal patterns and avoiding excessive masking that could disrupt contextual understanding.\footnote{A 30\% masking rate consistently delivered the best performance in most cases with BiTimeBERT. Our experiments on BiTimeBERT 2.0 further confirm that this setting also ensures strong performance across a range of downstream tasks, while maintaining methodological consistency with our prior work for ease of comparison.} 
Subsequently, we continue to randomly sample other tokens that are not part of $T$ or $S$, until a total of 15\% of the tokens in the sequence are sampled. For instance, in Figure~\ref{fig1}, tokens like "born" and "figure" are sampled and masked, whereas the temporal signal "during" and the temporal expression "2006" are not allowed to be sampled. Finally, the sampled tokens are treated as follows: 80\% of the sampled tokens are replaced with "[MASK]", 10\% are replaced with random tokens, and 10\% remain unchanged. For example, in the more detailed input shown in Figure~\ref{fig2}, the token "clear" is sampled but remains unchanged, "music" is sampled and replaced with the random token "next", and other sampled tokens are replaced with "[MASK]". Note also that the 15\% total token masking rate and the 80-10-10 replacement strategy are widely validated practices in numerous pre-trained language models like BERT, shown to balance token predictability and context learning effectively.

Through the ETAMLM masking scheme, BiTimeBERT 2.0 explicitly incorporates both the inherent domain knowledge embedded in news articles and the explicit temporal information, encompassing content time and temporal signals. This strategy fosters the model's ability to integrate knowledge about interrelated events and understand the connections among various temporal expressions. Specifically, this approach forces the model to consider the contextual relationships of surrounding, unmasked temporal details when predicting tokens for masked temporal expressions or signals. For instance, as illustrated in Figure~\ref{fig2}, the model is presented with a text describing the career of Mr. Smith, a notable figure in rap music, in which certain temporal expressions are masked. The text indicates that his influential career spanned "during [MASK] [MASK]" and that a pivotal shift in his stance occurred "[MASK] 2006". To accurately reconstruct these masked tokens, the model must infer from the available context that Mr. Smith's major career phase likely occurred in the 1990s and that his notable shift on violence and criminality emerged before 2006. This example clearly demonstrates the effectiveness of the ETAMLM scheme in encouraging the model to utilize both context and surrounding temporal cues. By requiring precise predictions of masked temporal elements, ETAMLM significantly enriches the model's understanding of temporal dynamics and enhances its overall temporal reasoning capabilities.

\subsection{Document Dating}
The second pre-training objective, Document Dating (DD), is specifically designed to incorporate timestamp information during its pre-training phase. In news archives, each article is typically annotated with a timestamp indicating its publication date. As discussed earlier, timestamps play an important role in numerous applications, particularly in temporal information retrieval \cite{li2003time, kanhabua2010determining, wang2021improving}. 

In the DD objective, a [CLS] token is inserted at the beginning of the input sequence, and its representation, $h_{[CLS]}$, serves as the contextual representation for the entire sequence. Unlike its conventional use for next sentence prediction in models like BERT, the [CLS] token here is tasked with predicting the document's timestamp, as shown in purple font in Figure~\ref{fig1}. An important hyper-parameter in this task is the granularity of the timestamp, which can vary from decade, year, month, to day temporal granularity. For instance, the timestamp of the news article in Figure~\ref{fig1} can be represented as "2007-05" under month granularity or simply as "2007" under year granularity. It is noteworthy that existing models like TempoBERT \cite{rosin2022time} typically adopt broader temporal granularities, such as years or decades, in their pre-training phase. While this broader granularity facilitates higher accuracy during pre-training by simplifying the prediction task, it often falls short in challenging time-related downstream tasks or in time-sensitive applications, such as event occurrence time prediction \cite{wang2021event} and web content analysis \cite{calzarossa2015modeling, mackay2017social}. 
In our prior research, systematic evaluations of different temporal granularities (year, month, and day) within the DD objective revealed that month granularity delivered the best performance in most cases. We observed that the DD objective using month granularity demonstrated significant performance improvements over that using year granularity in both year-based and month-based time prediction tasks. Moreoever, employing DD with day granularity often leads to relatively poorer performance, likely due to the increased complexity involved in predicting timestamps at this finer level of granularity. For instance, in the New York Times Annotated Corpus used in our experiments, which spans from 1987/01/01 to 2007/06/19, the number of classes for document dating under day granularity is a staggering 7,475, compared to just 246 under month granularity. This substantial increase in the number of classes necessitates a significantly larger pre-training dataset to achieve effective learning. Consequently, we continue to utilize month granularity for the DD objective in this work.

The DD objective, focused on predicting timestamp information, significantly enhances the model by infusing it with time-related and task-specific knowledge during pre-training. This pre-training strategy not only deepens the model's understanding of temporal contexts but also improves its proficiency in handling time-sensitive tasks, particularly in scenarios with small datasets that are typically insufficient for training task-agnostic language models effectively. By integrating timestamp prediction into the pre-training phase, the model gains a foundational capability to discern temporal patterns and relationships, which directly translates to improved performance on downstream time-related tasks. Similar ideas have been extensively employed in other studies, where language models are adapted to task-specific knowledge through tailored task-oriented pre-training objectives, leading to notable performance improvements. For example, \citet{xu2019bert} demonstrated the effectiveness of continually pre-training BERT with a reading comprehension objective, achieving superior results in review comprehension task. Likewise, \citet{ke2019sentilare} propose the Sentilare model, which incorporates sentence sentiment classification into the pre-training process, yielding strong performance on sentiment analysis task. Additionally, \citet{han2021fine} employ MLM alongside the proposed utterance relevance classification objective during pre-training, achieving improved results in response selection task.

\subsection{Time-sensitive Entity Replacement}

The third and final pre-training objective of BiTimeBERT 2.0, Time-sensitive Entity Replacement (TSER), focuses on dynamic entities and represents a significant advancement over our previous work \cite{wang2023bitimebert}. This objective draws inspiration from prior studies \cite{TNER, xu2022time, gonzalez2023injecting, agarwal2018dianed}, which highlight the close relationship between named entities and specific time periods, showing that named entities inherently carry temporal significance, particularly in the case of "Person" entities. For example, while incorporating temporal information into NER \cite{TNER} and NED \cite{agarwal2018dianed} tasks has been shown to enhance performance, "Person" entities consistently exhibit the most notable improvements, significantly outperforming other entity types such as "Location" and "Organization" by significant margins. These improvements can be attributed to the relatively shorter life spans or activity durations of "Person" entities, which make them inherently more time-sensitive compared to other named entity types. Building upon these insights, the TSER objective in BiTimeBERT 2.0 specifically targets "Person" entities due to their pronounced temporal sensitivity. 

In the TSER objective, we first construct a set of "Person" entities for each month within the pre-training corpus, comprising individuals mentioned in documents published during that specific month.\footnote{The New York Times Annotated Corpus spans 246 months from January 1987 to June 2007, yielding 246 distinct monthly "Person" entity sets, comprising approximately 3.4 million unique "Person" entities in total. Note that these entities are identified using spaCy during the same process that extracts temporal expressions for the ETAMLM task. } TSER specifically targets "Person" entities that remain unsampled after the ETAMLM procedure. For example, in Figure~\ref{fig2}, entities such as "Smith" and "Tupac Shakur" are considered, whereas "Notorious B.I.G." is excluded since its tokens are partially masked during ETAMLM. 
For the unsampled "Person" entities, TSER applies replacement to 50\% of them, substituting the original entity with another randomly selected "Person" entity from the same month's set. In the other 50\%, the entity remains unchanged. For example, as depicted in Figure~\ref{fig2}, "Tupac Shakur" is replaced with "Antonin Scalia" from the same month's set, while "Smith" is left unchanged. Finally, the model utilizes the concatenated boundary word representations of each "Person" entity to perform a binary classification task, predicting whether the entity has been "replaced" or "not replaced".

By incorporating the TSER objective, we aim to enhance the model's ability to analyze and interpret the temporal nuances associated with dynamic entities, particularly "Person" entities. This objective not only encourages the model to capture the implicit temporal associations of named entities but also strengthens its capacity to discern shifts in context caused by entity replacement, thereby improving its reasoning about time-sensitive information. Accurate identification and contextualization of "Person" entities within their specific temporal frames enables BiTimeBERT 2.0 to provide deeper insights into how historical events and cultural developments are framed in news texts. For example, a mention of "Barack Obama" in the context of U.S. politics is likely to refer to the period between 2008 and 2016, corresponding to his presidency, while a reference to "Greta Thunberg" in environmental activism would point to 2018 or later, coinciding with her rise as a global climate advocate. 
Through the TSER objective, BiTimeBERT 2.0 is trained to capture the implicit temporal information inherent in these entities, yielding richer insights into the interplay between individuals and their historical or cultural contexts, thereby enabling more nuanced and time-aware analyses of texts.

\section{Experimental Settings}
\subsection{Refined Pre-training Corpus Construction and BiTimeBERT 2.0 Implementation}
\label{section4_1}

In this study, we utilize the New York Times Annotated Corpus (NYT corpus) \cite{sandhaus2008new}, a well-established temporal news collection comprising over 1.8 million articles published between January 1987 and June 2007. The NYT corpus has been widely adopted in Temporal Information Retrieval research \cite{campos2015survey, DBLP:journals/ftir/KanhabuaBN15}, and our previous work \cite{wang2023bitimebert} demonstrated its effectiveness in generating time-aware language representations. 
To further optimize training efficiency and model effectiveness, we introduce a refinement process that excludes sentences lacking explicit content time. Articles devoid of any temporal references are omitted, resulting in a more focused dataset of approximately 1.4 million articles, referred to as the Refined NYT (RNYT) corpus. This refinement substantially reduces computational costs, cutting GPU hours from 80 in the original BiTimeBERT to just 38 in BiTimeBERT 2.0 under the same training configuration on a single NVIDIA A100 GPU, a reduction of nearly 53\%. Remarkably, despite training on a smaller corpora, BiTimeBERT 2.0 achieves comparable or superior performance in most cases, underscoring the effectiveness of focusing on temporally relevant content. This optimization is particularly crucial as modern AI research continues to scale up model sizes, making computational efficiency an increasingly critical consideration. 
Additionally, before pre-processing the corpus, 50,000 original articles were randomly sampled from the NYT corpus. This subset is exclusively designated for experiments on the document dating downstream task, as detailed in Section \ref{section4_2}. To ensure the validity of our experiments, these articles are excluded from the RNYT corpus during the pre-training phase.

To ensure a fair comparison with our prior work, we adopt the same pre-training hyperparameters used in BiTimeBERT, which were validated through empirical testing to ensure computational efficiency and model effectiveness. Specifically, we adopt BERT \cite{devlin2018bert} as the base framework,\footnote{Note that our method is applicable to all Transformer encoder-based language models.} initializing BiTimeBERT 2.0 with the pre-trained parameters of the $BERT_{BASE}$ (cased) model. This initialization serves as a robust starting point, mitigating the high computational costs associated with training from scratch. BiTimeBERT 2.0 is then continually pre-trained on the RNYT corpus for 10 epochs using the ETAMLM, DD, and TSER objectives. 
The selection of 10 epochs reflects a balance between achieving sufficient convergence for pre-training objectives and maintaining computational efficiency, as determined in our prior work through empirical testing based on downstream task performance. To handle the potentially long text in news articles, we set the maximum sequence length to 512 tokens, aligning with the inherent limit of the BERT model. Additionally, we employ a batch size of 8 with gradient accumulation steps set to 8, effectively balancing GPU memory constraints and training efficiency. Optimization is conducted using the AdamW optimizer \cite{kingma2014adam} with a learning rate of 3e-5, ensuring stable and effective training throughout the pre-training process.

\subsection{Downstream Tasks}
\label{section4_2}

We rigorously evaluate BiTimeBERT 2.0 across six datasets, spanning three distinct time-related downstream tasks: \emph{event occurrence time estimation}, \emph{document dating}, and \emph{semantic change detection}. These tasks hold significant implications for various domains within NLP and IR, such as improving the accuracy of historical event timelines \cite{steen2019abstractive, campos2021automatic, yu2021multi}, enhancing the precision of content retrieval based on specific time frames \cite{wang2020answering}, and tracking the evolution of language and meanings in digital archives \cite{wevers2020digital, tahmasebi2010terminology}. Each task is carefully designed to evaluate a specific aspect of the model's temporal understanding capabilities. For the task of event occurrence time estimation, we utilize the EventTime dataset \cite{wang2021event} and the WOTD dataset \cite{honovich2020machine}. This task focuses on accurately predicting the occurred time of specific events. In the document dating task, aimed at predicting the publication timestamps of documents, we employ the NYT-Timestamp dataset and the TDA-Timestamp dataset. Finally, the semantic change detection task assesses whether and to what extent the meanings of a set of target words have evolved over time. For this task, we leverage the LiverpoolFC dataset \cite{del2018short} and the SemEval-English dataset \cite{schlechtweg2020semeval}, both of which are time-annotated corpora specifically curated for studying semantic evolution. To determine how well a model can detect changes in the meaning of words over time, we measure its performance by comparing the model's assessment of semantic shift for each target word to the semantic index (i.e., the ground truth). These tasks collectively enable a comprehensive evaluation of BiTimeBERT 2.0's temporal understanding and its effectiveness across diverse time-sensitive applications.

The details of the six datasets are discussed below:
\begin{enumerate}[wide, labelwidth=!, labelindent=3pt]
\item \textbf{EventTime} \cite{wang2021event}: This dataset comprises 22,398 events with descriptions and occurrence times from January 1987 to June 2007, sourced from Wikipedia year pages\footnote{\url{https://en.wikipedia.org/wiki/List_of_years}} and "On This Day" website.\footnote{\url{https://www.onthisday.com/dates-by-year.php}} Since the compared SOTA method \cite{wang2021event} performs searches on the entire NYT corpus, we introduce an additional dataset, \textbf{EventTime-WithRelDoc}, designed to simulate a comparable input setting. Following the retrieval methodology outlined in \cite{wang2021event}, we employ an unsupervised keyword extraction method called Yake! \cite{campos2020yake} to identify keywords for each event description. The extracted keywords are then used as queries to retrieve relevant news articles from the NYT corpus based on the BM25 ranking. For each event, the top-1 relevant retrieved document is selected, and the new model input is constructed by combining the target event description with the appended timestamp and text content of the top-1 document.

\item \textbf{WOTD} \cite{honovich2020machine}: This dataset was scraped from Wikipedia's "On this day" webpages,\footnote{\url{https://en.wikipedia.org/wiki/Wikipedia:On_this_day/Today}.} comprising 6,809 short event descriptions along with their occurrence year information. WOTD encompasses 635 classes, each corresponding to a distinct occurrence year. The earliest year is 1302, the latest is 2018, with a median year of 1855.0 and a mean year of 1818.7. Additionally, the dataset provides contextual information (CI) for each event, consisting of relevant sentences extracted from Wikipedia through a series of carefully designed filtering steps, such as key entity extraction and sentence filtering.\footnote{For instance, the contextual information of the WOTD example in Table~\ref{tab_examples} contains sentence like "French forces were rebuilt, and feeling bitter about having lost many of France's overseas colonies to the British Empire during the Seven Years' War, Louis XVI was eager to give the American rebels financial and military support."} 
During evaluation, the dataset is tested both without contextual information and with contextual information, with the latter referred to as \textbf{WOTD-WithCI}. 
It is important to note that only year information is provided as gold labels, limiting the models to predict time at the year granularity. Furthermore, the time span of the WOTD dataset (1302-2018) significantly exceeds that of the NYT corpus (1987-2007) used for pre-training. This allows us to assess the models' robustness when confronted with a dataset spanning a much longer and older time range.

\item \textbf{NYT-Timestamp}: The NYT-Timestamp dataset is employed to evaluate the models on the document dating task. It comprises 50,000 news articles sampled separately from the NYT corpus \cite{sandhaus2008new}, as described in Section \ref{section4_1}.

\item \textbf{TDA-Timestamp\footnote{\url{https://www.gale.com/binaries/content/assets/gale-us-en/primary-sources/intl-gps/ghn-factsheets-fy18/ghn_factsheet_fy18_website_tda.pdf}}}: To further assess document dating, we also use the Times Digital Archive (TDA), which contains over 12 million news articles published across more than 200 years (1785-2012).\footnote{Note that despite TDA containing more articles and spanning a longer time period, the high prevalence of OCR errors in TDA led us to use it only for testing, not for pre-training. In comparison with the NYT corpus, errors are relatively common in TDA, especially in the early years \cite{niklas2010unsupervised}. For example, the average error rate from 1785 to 1932 was found to be above 30\%, with the highest rate reaching about 60\%.} 
Similar to the NYT-Timestamp dataset, we randomly sample 50,000 articles from the TDA corpus, with timestamps ranging from "1785/01/10" to "2009-12-31". Similar to the WOTD dataset, this extended time span enables a robust evaluation of the models' ability to handle long-term temporal contexts, further testing their resilience and effectiveness across diverse time periods.

\item \textbf{LiverpoolFC} \cite{del2018short}: The LiverpoolFC corpus, composed of Reddit posts from the Liverpool Football Club subreddit, is tailored for short-term meaning shift analysis in online communities. The data spans two distinct periods: 2011-2013 and 2017. In our study, we apply minimal text pre-processing, primarily removing URLs, following the methodology outlined by \citet{rosin2022time}. The semantic change evaluation utilizes a set of 97 words from this corpus, each annotated with semantic shift labels by the members of the LiverpoolFC subreddit \cite{del2018short}. These labels range from 0 to 1, representing the degree of semantic change, with the average judgment serving as the gold standard semantic shift index.

\item \textbf{SemEval-English} \cite{schlechtweg2020semeval}: In contrast to the LiverpoolFC dataset, the SemEval-English dataset from SemEval-2020 Task 1 focuses on long-term semantic change detection, covering two centuries. SemEval-English dataset spans two distinct periods: 1810-1860 and 1960-2010. This task comprises two subtasks: (1) binary classification of whether a word sense has been gained or lost, and (2) ranking a list of words based on their degree of semantic change. For this study, we utilize the dataset for subtask 2, consisting of 37 target words annotated with graded labels indicating  the extent of semantic shift. These target words are carefully balanced in terms of part of speech (POS) and frequency. Following prior work \cite{rosin2022time}, POS tags are omitted from both the corpus and the evaluation set. 
\end{enumerate}

\begin{table}[]
\caption{Statistics of six datasets across three distinct downstream tasks.}
\label{tab_datasets}
\begin{subfigure}{0.99\textwidth}
\centering
\footnotesize
\vspace{-1.0em}
\renewcommand{\arraystretch}{1.2}
\begin{tabular}{|c|c|c|c|c|c|}
\hline
\textbf{Dataset} & \textbf{Size} & \textbf{Time Span} & \textbf{Source} & \textbf{Granularity} & \textbf{Task} \\ \hline
EventTime & 22,398 & 1987-2007 & \begin{tabular}[c]{@{}c@{}}Wikipedia \& \\ "On This Day" Website\end{tabular} & Year, Month, Day & \begin{tabular}[c]{@{}c@{}}Event Occurrence \\ Time Estimation\end{tabular} \\ \hline
WOTD & 6,809 & 1302-2018 & Wikipedia Website & Year & \begin{tabular}[c]{@{}c@{}}Event Occurrence \\ Time Estimation\end{tabular} \\ \hline
NYT-Timestamp & 50,000 & 1987-2007 & News Archive & Year, Month, Day & Document Dating \\ \hline
TDA-Timestamp & 50,000 & 1785-2009 & News Archive & Year, Month, Day & Document Dating \\ \hline
\end{tabular}
\vspace{0.2em}
\caption{Statistics of event occurrence time estimation datasets and document dating datasets.}
\label{tab_datasets1}
\end{subfigure}

\vspace{1.0em}
\begin{subfigure}{0.99\textwidth}
\centering
\footnotesize
  
  \vspace{-1.0em}
  \label{tab_scd_datasets}
  \renewcommand{\arraystretch}{1.3}
   \begin{tabular}{|c|c|c|c|c|c|}
  \hline
  \makecell{\textbf{Dataset}} & \textbf{Target Words}& \textbf{ C1 Source} & \textbf{C1 Time Period} &\textbf{C2 Source} & \textbf{C2 Time Period} \\
    \hline
   LiverpoolFC  & 97 &  Reddit &   2011–2013 & Reddit & 2017 \\
    \hline
    SemEval-English  & 37 &  CCOHA &   1810–1860 & CCOHA & 1960–2010 \\
    \hline
\end{tabular}
\vspace{0.2em}
\caption{Statistics of semantic change detection datasets.}
\label{tab_datasets2}
\end{subfigure}
\vspace{-2.0em}
\end{table}

\begin{table}[]
\caption{Sample data from six datasets across three distinct downstream tasks.}
\vspace{-1.0em}
\label{tab_examples}
\begin{subfigure}{0.99\textwidth}
\centering
\footnotesize
\renewcommand{\arraystretch}{1.3}

\begin{tabular}{|c|l|c|}
\hline
\textbf{Dataset} & \multicolumn{1}{c|}{\textbf{Text (Event Description or Document Content)}} & \textbf{Time} \\ \hline
EventTime & \begin{tabular}[c]{@{}l@{}}Mary Kay LeTourneau, 36, former teacher, who violated probation \\ by seeing 14 year old father of her baby, sentenced to 7 years.\end{tabular} & 2006-09 \\ \hline
WOTD & \begin{tabular}[c]{@{}l@{}}Louis XVI of France gives his public assent to Civil Constitution of \\ the Clergy during the French Revolution.\end{tabular} & 1790 \\ \hline
NYT-Timestamp & \begin{tabular}[c]{@{}l@{}}NASA managers said Wednesday that they would probably ''never \\ know for sure'' what caused unexpected and mysterious debris to \\ appear around the space shuttle Atlantis in the past two days ...\end{tabular} & 2006-09 \\ \hline
TDA-Timestamp & \begin{tabular}[c]{@{}l@{}}Sir,-When a guest some 50 rears ago at an indigo factory in Bihar, I \\ remember two Moslems paving mv host a visit accompanied by a \textbackslash{}" \\ wolf-child ' that was on a lead and progressed on her arms from ...\end{tabular} & 1927-04 \\ \hline
\end{tabular}
\vspace{0.2em}
\caption{Sample data from event occurrence time estimation datasets and document dating datasets.}
\label{tab_examples1}
\end{subfigure}

\vspace{-0.2em}

\begin{subfigure}{0.99\textwidth}
\renewcommand{\arraystretch}{1.3}
\centering
\footnotesize
\begin{tabular}{|c|c|}
\hline
\textbf{Dataset} & \textbf{Target Words with their Corresponding Shift Indices} \\ \hline
LiverpoolFC & toxic (0.857); f5 (0.556); arrangement (0.413); vomit (0) ... \\ \hline
SemEval-English & plane (0.882); tip (0.678); attack (0.143); chairman (0) ... \\ \hline
\end{tabular}
\vspace{0.2em}
\caption{Sample data from semantic change detection datasets.}
\label{tab_examples2}
\end{subfigure}
\vspace{-2.0em}
\end{table}

Table~\ref{tab_datasets} provides a summary of the basic statistics for the six datasets, organized into two subtables for ease of reference. Table~\ref{tab_datasets1} outlines the four datasets utilized for event occurrence time estimation and document dating tasks, while Table~\ref{tab_datasets2} focuses on the two datasets employed for semantic change detection. Notably, the WOTD and TDA-Timestamp datasets span significantly longer time periods compared to the pre-training NYT corpus. Similarly, the SemEval-English dataset spans across two centuries, extending beyond the NYT corpus timeframe. The LiverpoolFC corpus, meanwhile, contains text from a timeframe not covered in the NYT corpus. 
Following prior research \cite{wang2021event, honovich2020machine, wang2023bitimebert}, the EventTime, WOTD, NYT-Timestamp, and TDA-Timestamp datasets are divided into training, validation, and testing sets using an 80:10:10 ratio. For the semantic change detection tasks, however, no additional data processing is applied to the LiverpoolFC and SemEval-English datasets, as per prior research \cite{rosin2022time, wang2023bitimebert}.

Table~\ref{tab_examples} presents sample data from each of the six datasets. As shown in Table~\ref{tab_examples1}, the datasets for EventTime, NYT-Timestamp, and TDA-Timestamp include detailed occurrence time or timestamp information. This allows the models tested on these datasets to be fine-tuned for estimating time with various temporal granularities (e.g., year or month). Conversely, models fine-tuned on WOTD are restricted to predicting time at a year granularity. The complexity of the task increases when estimating time at finer granularities, such as months, due to the substantial increase in the number of possible labels. For instance, the TDA-Timestamp dataset, when considered at month granularity, comprises 2,688 distinct labels corresponding to the total number of months covered. Meanwhile, Table~\ref{tab_examples2} offers examples of target words and their corresponding semantic shift indices from the LiverpoolFC and SemEval-English datasets, illustrating the extent of semantic change over time. For example, the target word "plane" in the SemEval-English dataset exhibits a high semantic shift index value (0.882), indicating a significant semantic change between 1810–1860 and 1960–2010. During the earlier period of 1810-1860, "plane" primarily referred to a flat surface, as powered airplanes had not yet been invented (the first flight occurred in 1903). By 1960–2010, however, the meaning of "plane" had expanded to include its now-common definition as a flying vehicle.

\subsection{Evaluation Metrics}
\begin{table}[]
\caption{Overview of the Pre-trained Language Models Used for Comparison.}
\label{tab_baselines}
\centering
\footnotesize
\vspace{-1.0em}
\renewcommand{\arraystretch}{1.2}
\begin{tabular}{|l|c|l|}
\hline
\multicolumn{1}{|c|}{\textbf{Language Models}} & \textbf{Pre-training Corpus} & \multicolumn{1}{c|}{\textbf{Pre-training Objectives}} \\ \hline
BERT-Orig                             & BooksCorpus + Wikipedia      & MLM, NSP                                              \\ \hline
BERT-NYT                              & NYT Corpus                   & MLM, NSP                                              \\ \hline
BERT-RNYT                             & RNYT Corpus                  & MLM, NSP                                              \\ \hline
BERT+TIR                              & NYT Corpus                   & MLM, TIR                                              \\ \hline
TempoBERT                             & Timestamped Texts            & MLM, TM                                               \\ \hline
BiTimeBERT                            & NYT Corpus                   & TAMLM, DD                                             \\ \hline
BiTimeBERT-RNYT                       & RNYT Corpus                  & TAMLM, DD                                             \\ \hline
BiTimeBERT+TSEMLM                     & RNYT Corpus                  & TAMLM, DD, TSEMLM                                     \\ \hline
BiTimeBERT+TRWR                       & RNYT Corpus                  & TAMLM, DD, TRWR                                       \\ \hline
BiTimeBERT 2.0                        & RNYT Corpus                  & ETAMLM, DD, TSER                                      \\ \hline
\end{tabular}
\end{table}

In our study, the first two downstream tasks, event occurrence time estimation and document dating, involve time prediction. For these tasks, we adopt accuracy (ACC) and mean absolute error (MAE) as evaluation metrics, following the approaches in \cite{wang2021event, wang2023bitimebert}. For the task of semantic shift detection, we evaluate performance by measuring the correlation between the ground truth semantic shift index and the model's assessment for each word in the evaluation set. To this end, we utilize both Pearson and Spearman correlation coefficients, aligning with the methodology in \cite{rosin2022time}. These metrics are explained as follows:

\begin{enumerate}[label=\arabic*),leftmargin=1.8em]
\item \textbf{Accuracy (ACC)}: This metric calculates the percentage of instances for which the model correctly predicts the actual time.
\item \textbf{Mean absolute error (MAE)}: MAE represents the average absolute difference between the model's predicted time and the actual time, tailored to the specified temporal granularity.\footnote{For example, an MAE of 1 at year granularity implies an average temporal distance error of 1 year, whereas at month granularity, it implies an average error of 1 month.}
\item \textbf{Pearson}: Pearson correlation assesses the linear correlation between the predicted semantic shifts and the ground truth. It evaluates how well the model's predictions reflect the actual magnitude of changes in word meanings.
\item \textbf{Spearman}: Unlike Pearson, Spearman correlation focuses on the rank order of predicted semantic shifts. It evaluates how well the relationship between predicted and actual semantic shifts can be described using a monotonic function, without assuming linearity.

\end{enumerate}

It is worth noting that, with the exception of the WOTD dataset which contains only year information, all models in our study could be evaluated under three different temporal granularities (day, month, and year), for both the event occurrence time estimation and document dating tasks. However, we observed that all pre-trained language models consistently performed poorly at the day granularity level, likely due to the significantly increased complexity and label sparsity at this level of granularity. Given this consistent underperformance, we have chosen to exclude results at day granularity from our analysis, focusing instead on month and year granularities, where the models demonstrate more reliable and meaningful performance.

\subsection{Compared Models}
\label{section4_4}

To evaluate the effectiveness of BiTimeBERT 2.0, we compare its performance against a range of baseline and state-of-the-art (SOTA) models across various time-related tasks. Table~\ref{tab_baselines} provides an overview of 10 pre-trained language models used for comparison, highlighting their differences in pre-training corpora and objectives. Below, we detail all the compared models, including both pre-trained and non-pre-trained approaches, used in this study:

\begin{enumerate}[wide, labelwidth=!, labelindent=3pt]
\item \textbf{RG (Random Guess)}: This baseline represents random guesses, with results computed as the average performance across 1,000 random selections. This serves as a lower-bound benchmark for the tasks of event occurrence time estimation and document dating.

\item \textbf{SOTA-EOTE}: These state-of-the-art methods for event occurrence time estimation are designed specifically for the EventTime and WOTD datasets, as proposed by \cite{wang2021event} and \cite{honovich2020machine}, respectively. Notably, these methods are not based on language models but instead rely on task-specific pipelines involving complex rules and multiple steps of searching and filtering to extract features for accurate date estimation. While these approaches achieve strong performance in their respective tasks, their reliance on handcrafted rules and tailored designs makes them less generalizable or easily applicable to other related tasks.

\item \textbf{BERT Variants}: 
\begin{itemize}[leftmargin=10mm]
    \item \textbf{BERT-Orig}: The original $BERT_{BASE}$ (cased) model \cite{devlin2018bert}, trained on the BooksCorpus \cite{zhu2015aligning} and English Wikipedia.
    \item \textbf{BERT-NYT}: A variant of BERT that is continually pre-trained on the full NYT corpus for 10 epochs, employing Masked Language Modelling (MLM) and Next Sentence Prediction (NSP) tasks.
    \item \textbf{BERT-RNYT}: Similar to BERT-NYT but pre-trained on the RNYT corpus, which excludes sentences lacking explicit temporal content. 
\end{itemize}

\item \textbf{BERT+TIR}: This model, introduced in our previous work \cite{wang2023bitimebert}, represents an alternative approach to leveraging content temporal information. BERT+TIR involves the continuous pre-training of BERT on the NYT corpus for 10 epochs, incorporating both MLM and TIR (temporal information replacement) tasks. The TIR task involves predicting whether temporal expressions in the text have been replaced or not. Specifically, in 50\% of cases, temporal expressions are substituted with others that maintain the same temporal granularity. This method aims to explore a different facet of using content time information.

\item \textbf{TempoBERT}: TempoBERT \cite{rosin2022time} is a time-aware BERT model that incorporates timestamp information through a pre-training objective named Time Masking (TM). This objective involves prepending texts with timestamps and masking them during training, as discussed in Section \ref{section2_4}. We include TempoBERT in comparisons for the semantic change detection task, where TempoBERT is primarily focused. \citet{rosin2022time} have also demonstrated that TempoBERT experiences a performance decline in time prediction tasks when compared to a simply fine-tuned BERT model, highlighting its limitations in tasks outside its main focus. 

\item \textbf{BiTimeBERT Variants}: 
\begin{itemize}[leftmargin=10mm]
\item \textbf{BiTimeBERT}: The original version of BiTimeBERT, introduced in our previous work \cite{wang2023bitimebert}. This model is pre-trained on the full NYT corpus for 10 epochs using the TAMLM and DD tasks.
\item \textbf{BiTimeBERT-RNYT}: A variant of BiTimeBERT that undergoes pre-training on the Refined NYT (RNYT) corpus, incorporating the same TAMLM and DD tasks. 

\item \textbf{BiTimeBERT+TSEMLM}: This variant introduces a new pre-training objective, \emph{time-sensitive entity masked language modeling} (TSEMLM), alongside TAMLM and DD tasks, and is pre-trained on the RNYT corpus. The TSEMLM task specifically masks "Person" entities within the text after masking content temporal information, and the model is then tasked with predicting these masked tokens. By masking 30\% of "Person" entities, TSEMLM enhances the model's sensitivity to the temporal dynamics associated with individuals. This additional focus on "Person" entities is implemented to explore whether an alternative approach of incorporating time-sensitive entity information can yield improved performance.

\item \textbf{BiTimeBERT+TRWR}: This variant introduces another alternative pre-training objective, \emph{temporal relation word replacement} (TRWR), alongside the TAMLM and DD tasks, and is also pre-trained on the RNYT corpus. TRWR targets temporal signals within the text, replacing them in 50\% of instances with alternatives of a different temporal nature (e.g., substituting "before" with "after" or "during"). The model is then tasked with determining whether the temporal signals in the text have been replaced, encouraging it to develop a nuanced understanding of altered temporal contexts. The primary goal of TRWR is to assess whether this strategy of temporal signal replacement can further improve the model's performance.

\item \textbf{BiTimeBERT 2.0}: Our latest version, BiTimeBERT 2.0, is continually pre-trained on the RNYT corpus and incorporates comprehensive enhancements across the ETAMLM, DD, and TSER tasks. Notably, this version includes refinements such as an enhanced emphasis on accurately interpreting temporal signals during the ETAMLM process.

\end{itemize}
\end{enumerate}

\subsection{Fine-tuning Setting}
For the event occurrence time prediction and document dating tasks, we fine-tune the aforementioned language models on the respective datasets. Each language model utilizes the final hidden state of the first token [CLS], denoted as $h_{[CLS]}$, as the representative vector for the entire sequence. A softmax classifier, parameterized by a weight matrix $X\in \mathbb{R}^{K x H}$, where $K$ is the number of classes in the dataset, is then appended. The models are optimized using the cross-entropy loss with the Adam optimizer. To account for the varying input lengths across different datasets, we set the maximum sequence length to 128 for the EventTime and WOTD datasets, which typically consist of shorter event descriptions. In contrast, for datasets with longer input sequences, such as EventTime-WithRelDoc, WOTD-WithCI, NYT-Timestamp, and TDA-Timestamp, we increase the maximum sequence length to 512. 
Hyperparameter tuning is performed via grid search over other key parameters, including batch size \(\{16, 32\}\), learning rate \(\{2\text{e-}5, 5\text{e-}5\}\), and epoch number \(\{5, 10, 15\}\). To ensure the reliability of our results, we run each experiment five times and utilize two-tailed t-tests to assess the statistical significance of performance differences between BiTimeBERT 2.0 and the baseline models.\footnote{For the main results presented in Table~\ref{tab_main1} to ~\ref{tab_main3}, the baseline for the two-tailed t-test is BiTimeBERT-RNYT. For the ablation studies in Table~\ref{tab_abl1} to ~\ref{tab_abl2}, the baseline for the two-tailed t-test is ETAMLM+DD.}

For the semantic change detection task, we first adapt the language models to their respective domains using either the LiverpoolFC corpus or the SemEval-English corpus, depending on the dataset in focus. To ensure a fair comparison with TempoBERT \cite{rosin2022time}, we adopt the same training hyperparameters. Specifically, for LiverpoolFC, the learning rate is set to 1e-7, and training is conducted for one epoch, while for SemEval-English, a learning rate of 1e-6 is used, with training spanning two epochs. To ensure uniformity in domain adaptation, all models are continually trained using Masked Language Modeling (MLM). It is important to note that due to the sentence-level nature of these corpora, which typically lack content time information, we do not apply ETAMLM in this phase. After the language models are trained and adapted to the specific domains, we follow TempoBERT's methodology to generate representations of target words for each time period. 
These representations are then used to assess the degree of semantic change for each word by calculating the cosine distance (cos\_dist) between representations. The model's performance is evaluated using Pearson's correlation coefficient and Spearman's rank correlation coefficient, which measure the alignment between predicted and ground truth semantic shifts. Additionally, we report results using the "time-diff" distance method, a metric tailored to TempoBERT, as introduced by \citet{rosin2022time}. To substantiate the reliability of the results, we again conduct statistical significance tests to assess whether the observed performance improvements over baseline models are statistically significant.

\section{Experimental Results}
\subsection{Main Results} 
\label{section5_1}

\subsubsection{Event Occurrence Time Estimation}
\label{section5_1_1}

Table~\ref{tab_main1} presents the performance of various models on event occurrence time estimation, evaluated using the EventTime, EventTime-WithRelDoc, WOTD, and WOTD-WithCI datasets. Compared to state-of-the-art event occurrence time estimation methods (SOTA-EOTE), BiTimeBERT 2.0 generally achieves superior results, with the exception of the ACC metric for EventTime-WithRelDoc under month granularity. Additionally, BiTimeBERT 2.0 consistently outperforms other models pre-trained on the RNYT corpus, including BiTimeBERT-RNYT, in terms of ACC and MAE across different settings such as year/month granularities, with/without top-1 relevant documents, and with/without contextual information. Notably, BiTimeBERT 2.0 demonstrates state-of-the-art performance on the EventTime-WithRelDoc dataset under year granularity and excels on the WOTD and WOTD-WithCI datasets, both of which span time periods significantly beyond its pre-training temporal range.

\begin{table}[]
\footnotesize
\renewcommand{\arraystretch}{1.3}
\caption{Performance comparison of different models on the event occurrence time estimation task, evaluated using the EventTime, EventTime-WithRelDoc, WOTD, and WOTD-WithCI datasets. Bold font indicates the best performing model while underline is used to mark the second best one. Asterisk (*) denotes statistical significance at \(p < 0.01\), and dagger (†) denotes significance at \(p < 0.001\).}
\vspace{-1.0em}
\label{tab_main1}
\begin{tabular}{|ccccccccccccc|}
\hline
\multicolumn{1}{|c|}{\multirow{3}{*}{\textbf{Model}}} & \multicolumn{4}{c|}{\textbf{EventTime}} & \multicolumn{4}{c|}{\textbf{EventTime-WithRelDoc}} & \multicolumn{2}{c|}{\textbf{WOTD}} & \multicolumn{2}{c|}{\textbf{WOTD-WithCI}} \\ \cline{2-13} 
\multicolumn{1}{|c|}{} & \multicolumn{2}{c|}{\textbf{Year}} & \multicolumn{2}{c|}{\textbf{Month}} & \multicolumn{2}{c|}{\textbf{Year}} & \multicolumn{2}{c|}{\textbf{Month}} & \multicolumn{1}{c|}{\multirow{2}{*}{\textbf{ACC}}} & \multicolumn{1}{c|}{\multirow{2}{*}{\textbf{MAE}}} & \multicolumn{1}{c|}{\multirow{2}{*}{\textbf{ACC}}} & \multirow{2}{*}{\textbf{MAE}} \\ \cline{2-9}
\multicolumn{1}{|c|}{} & \multicolumn{1}{c|}{\textbf{ACC}} & \multicolumn{1}{c|}{\textbf{MAE}} & \multicolumn{1}{c|}{\textbf{ACC}} & \multicolumn{1}{c|}{\textbf{MAE}} & \multicolumn{1}{c|}{\textbf{ACC}} & \multicolumn{1}{c|}{\textbf{MAE}} & \multicolumn{1}{c|}{\textbf{ACC}} & \multicolumn{1}{c|}{\textbf{MAE}} & \multicolumn{1}{c|}{} & \multicolumn{1}{c|}{} & \multicolumn{1}{c|}{} &  \\ \hline
\multicolumn{1}{|c|}{RG} & \multicolumn{1}{c|}{4.77} & \multicolumn{1}{c|}{6.92} & \multicolumn{1}{c|}{0.41} & \multicolumn{1}{c|}{81.60} & \multicolumn{1}{c|}{4.77} & \multicolumn{1}{c|}{6.92} & \multicolumn{1}{c|}{0.40} & \multicolumn{1}{c|}{81.70} & \multicolumn{1}{c|}{0.16} & \multicolumn{1}{c|}{217.72} & \multicolumn{1}{c|}{0.15} & 217.57 \\ \hline
\multicolumn{1}{|c|}{SOTA-EOTE} & \multicolumn{1}{c|}{-} & \multicolumn{1}{c|}{-} & \multicolumn{1}{c|}{-} & \multicolumn{1}{c|}{-} & \multicolumn{1}{c|}{40.93} & \multicolumn{1}{c|}{3.01} & \multicolumn{1}{c|}{\textbf{30.89}} & \multicolumn{1}{c|}{36.19} & \multicolumn{1}{c|}{11.40} & \multicolumn{1}{c|}{-} & \multicolumn{1}{c|}{13.10} & - \\ \hline
\multicolumn{1}{|c|}{BERT-Orig} & \multicolumn{1}{c|}{21.65} & \multicolumn{1}{c|}{3.47} & \multicolumn{1}{c|}{5.09} & \multicolumn{1}{c|}{43.81} & \multicolumn{1}{c|}{35.98} & \multicolumn{1}{c|}{3.89} & \multicolumn{1}{c|}{5.98} & \multicolumn{1}{c|}{37.95} & \multicolumn{1}{c|}{7.20} & \multicolumn{1}{c|}{52.58} & \multicolumn{1}{c|}{9.69} & 41.16 \\ \hline
\multicolumn{13}{|c|}{\textit{Models Pre-trained using the entire NYT corpus}} \\ \hline
\multicolumn{1}{|c|}{BERT-NYT} & \multicolumn{1}{c|}{21.25} & \multicolumn{1}{c|}{3.56} & \multicolumn{1}{c|}{5.18} & \multicolumn{1}{c|}{43.50} & \multicolumn{1}{c|}{34.46} & \multicolumn{1}{c|}{4.45} & \multicolumn{1}{c|}{8.21} & \multicolumn{1}{c|}{34.14} & \multicolumn{1}{c|}{8.08} & \multicolumn{1}{c|}{53.75} & \multicolumn{1}{c|}{19.97} & 36.47 \\ \hline
\multicolumn{1}{|c|}{BERT+TIR} & \multicolumn{1}{c|}{25.40} & \multicolumn{1}{c|}{3.23} & \multicolumn{1}{c|}{6.83} & \multicolumn{1}{c|}{40.45} & \multicolumn{1}{c|}{36.47} & \multicolumn{1}{c|}{3.54} & \multicolumn{1}{c|}{17.01} & \multicolumn{1}{c|}{31.72} & \multicolumn{1}{c|}{10.13} & \multicolumn{1}{c|}{54.92} & \multicolumn{1}{c|}{18.36} & 35.99 \\ \hline
\multicolumn{1}{|c|}{BiTimeBERT} & \multicolumn{1}{c|}{\textbf{31.91}} & \multicolumn{1}{c|}{\textbf{3.12}} & \multicolumn{1}{c|}{\textbf{12.99}} & \multicolumn{1}{c|}{\textbf{34.79}} & \multicolumn{1}{c|}{{\ul 41.96}} & \multicolumn{1}{c|}{2.40} & \multicolumn{1}{c|}{{\ul 25.76}} & \multicolumn{1}{c|}{\textbf{28.86}} & \multicolumn{1}{c|}{{\ul 11.60}} & \multicolumn{1}{c|}{{\ul 48.51}} & \multicolumn{1}{c|}{{\ul 23.05}} & 33.70 \\ \hline
\multicolumn{13}{|c|}{\textit{Models Pre-trained using the RNYT corpus}} \\ \hline
\multicolumn{1}{|c|}{BERT-RNYT} & \multicolumn{1}{c|}{22.84} & \multicolumn{1}{c|}{3.51} & \multicolumn{1}{c|}{5.82} & \multicolumn{1}{c|}{42.48} & \multicolumn{1}{c|}{34.81} & \multicolumn{1}{c|}{2.75} & \multicolumn{1}{c|}{13.15} & \multicolumn{1}{c|}{34.72} & \multicolumn{1}{c|}{8.08} & \multicolumn{1}{c|}{48.63} & \multicolumn{1}{c|}{19.44} & 33.17 \\ \hline
\multicolumn{1}{|c|}{BiTimeBERT-RNYT} & \multicolumn{1}{c|}{28.36} & \multicolumn{1}{c|}{3.27} & \multicolumn{1}{c|}{8.93} & \multicolumn{1}{c|}{38.52} & \multicolumn{1}{c|}{39.21} & \multicolumn{1}{c|}{2.47} & \multicolumn{1}{c|}{20.15} & \multicolumn{1}{c|}{30.53} & \multicolumn{1}{c|}{10.30} & \multicolumn{1}{c|}{52.37} & \multicolumn{1}{c|}{21.68} & 34.29 \\ \hline
\multicolumn{1}{|c|}{BiTimeBERT+TSEMLM} & \multicolumn{1}{c|}{28.27} & \multicolumn{1}{c|}{3.24} & \multicolumn{1}{c|}{9.41} & \multicolumn{1}{c|}{39.84} & \multicolumn{1}{c|}{38.82} & \multicolumn{1}{c|}{2.55} & \multicolumn{1}{c|}{18.37} & \multicolumn{1}{c|}{31.60} & \multicolumn{1}{c|}{9.89} & \multicolumn{1}{c|}{57.15} & \multicolumn{1}{c|}{20.73} & 36.21 \\ \hline
\multicolumn{1}{|c|}{BiTimeBERT+TRWR} & \multicolumn{1}{c|}{27.60} & \multicolumn{1}{c|}{3.20} & \multicolumn{1}{c|}{9.18} & \multicolumn{1}{c|}{38.76} & \multicolumn{1}{c|}{40.35} & \multicolumn{1}{c|}{{\ul 2.30}} & \multicolumn{1}{c|}{19.89} & \multicolumn{1}{c|}{30.12} & \multicolumn{1}{c|}{10.37} & \multicolumn{1}{c|}{53.64} & \multicolumn{1}{c|}{22.25} & {\ul 31.31} \\ \hline
\multicolumn{1}{|c|}{BiTimeBERT 2.0} & \multicolumn{1}{c|}{{\ul 29.43}$^{\dagger}$} & \multicolumn{1}{c|}{{\ul 3.17}$^{*}$} & \multicolumn{1}{c|}{{\ul 10.62}$^{\dagger}$} & \multicolumn{1}{c|}{{\ul 36.15}$^{\dagger}$} & \multicolumn{1}{c|}{\textbf{42.07}$^{\dagger}$} & \multicolumn{1}{c|}{\textbf{2.23}$^{*}$} & \multicolumn{1}{c|}{20.25$^{*}$} & \multicolumn{1}{c|}{{\ul 30.07}$^{\dagger}$} & \multicolumn{1}{c|}{\textbf{12.42}$^{\dagger}$} & \multicolumn{1}{c|}{\textbf{46.22}$^{\dagger}$} & \multicolumn{1}{c|}{\textbf{24.76}$^{\dagger}$} & \textbf{30.42}$^{*}$ \\ \hline
\end{tabular}

\end{table}

On the \textbf{EventTime} dataset, BiTimeBERT 2.0 demonstrates impressive gains over BERT-RNYT, with improvements ranging from 28.85\% to 82.47\% in ACC and from 9.68\% to 14.90\% in MAE across year and month granularities. The original BiTimeBERT \cite{wang2023bitimebert}, trained with the TAMLM and DD objectives using the full NYT corpus, maintains a strong performance, validating the effectiveness of these objectives in capturing detailed temporal information from the news collection. While BiTimeBERT 2.0 does not fully match the performance of BiTimeBERT on the EventTime dataset, it still achieves commendable second-best results using a significantly smaller pre-training corpus and dramatically reduced pre-training time. This efficiency is particularly critical given the increasing importance of resource optimization in the development of large-scale AI models. 
For the \textbf{EventTime-WithRelDoc} dataset, where the top-1 relevant document is considered, BiTimeBERT 2.0 exhibits remarkable improvements over BERT-NYT, which is trained on the entire NYT corpus. At the month granularity, BiTimeBERT 2.0 achieves a 146.65\% improvement in ACC and an 11.92\% reduction in MAE. Its superiority over BERT-Orig is even more pronounced, with improvements of 238.62\% in ACC and 20.76\% in MAE at the month granularity.
It is worth noting that the SOTA-EOTE method \cite{wang2021event} for the EventTime-WithRelDoc dataset involves a time-consuming preparation process for time estimation. This approach relies on constructing multivariate time series from the top-50 retrieved documents, employing complex steps such as sentence similarity computation. In contrast, our method utilizes the simpler top-1 relevant document ranked by BM25, making it significantly more efficient. We believe that the performance of BiTimeBERT 2.0 could be further improved by incorporating more relevant information through advanced information retrieval techniques. Finally, in comparison with the original BiTimeBERT, BiTimeBERT 2.0 excels particularly under year granularity, while being less effective in the month granularity.

On the \textbf{WOTD} dataset, BiTimeBERT 2.0 consistently achieves the best results across both ACC and MAE metrics. For instance, it achieves a 20.58\% improvement in ACC and an 11.74\% reduction in MAE compared to BiTimeBERT-RNYT, while also outperforming the original BiTimeBERT by 7.06\% in ACC and 4.72\% in MAE. On the \textbf{WOTD-WithCI} dataset, where contextual information is incorporated, BiTimeBERT 2.0 demonstrates even greater gains. It achieves a 14.20\% increase in ACC and a 11.28\% reduction in MAE compared to BiTimeBERT-RNYT, and surpasses the original BiTimeBERT model with a 7.41\% ACC improvement and a 9.73\% reduction in MAE.\footnote{As discussed in Section \ref{section4_2}, contextual information includes relevant sentences extracted from Wikipedia, providing additional external knowledge.} The exceptional performance of BiTimeBERT 2.0 on these two datasets, which span temporal ranges significantly beyond its pre-training corpus, highlights its robust generalization capabilities. These results underscore the model's ability to effectively integrate temporal reasoning and leverage external contextual information, even when handling datasets with vastly different and extended temporal ranges.

Additionally, the BERT-Orig model yields poor results compared to the other BERT variants (BERT-NYT and BERT-RNYT). Interestingly, BERT-RNYT, trained on a substantially smaller corpus, outperforms BERT-NYT in most cases. This suggests that for temporal news collections, a more focused and refined dataset like the RNYT corpus may lead to more effective results when applying standard BERT objectives. Furthermore, the BiTimeBERT variants incorporating the proposed TSEMLM or TRWR objectives also show strong performance. Between the two variants, BiTimeBERT+TRWR consistently outperforms BiTimeBERT+TSEMLM, except on the EventTime dataset. However, both of these newly proposed BiTimeBERT variants significantly outperform the other three BERT variants (BERT-Orig, BERT-NYT, and BERT-RNYT) by large margins in most cases. These results underscore the advantages of integrating time-sensitive entities and temporal signals into the training process, highlighting their potential to enhance the model's temporal reasoning capabilities and overall effectiveness in time-sensitive tasks.

\subsubsection{Document Dating}

Table~\ref{tab_main2} presents the results of the document dating task on the \textbf{NYT-Timestamp} and \textbf{TDA-Timestamp} datasets. Similarly, BiTimeBERT 2.0 consistently outperforms other language models pre-trained on the RNYT corpus across both datasets. Specifically, compared to BiTimeBERT-RNYT, BiTimeBERT 2.0 achieves ACC improvements ranging from 8.63\% to 23.80\% on the NYT-Timestamp dataset across year and month granularities, and from 20.14\% to 128.03\% on the TDA-Timestamp dataset. These results further underscore the effectiveness of the pre-training objectives employed in BiTimeBERT 2.0. 
However, when compared to the original BiTimeBERT, BiTimeBERT outperforms BiTimeBERT 2.0 on the NYT-Timestamp dataset. This is likely due to BiTimeBERT’s pre-training on the significantly larger, unprocessed NYT corpus, which may include articles with content and publication dates closely aligned to those in the NYT-Timestamp dataset. For instance, events reported in the NYT-Timestamp dataset could overlap with similar articles in the pre-training NYT corpus, as the same event might be covered in multiple news articles over a short time frame. For example, a major event like an earthquake might generate a series of related articles spanning several days, discussing its immediate impact, aftermath, and developments. Such overlap could introduce some degree of information leakage during pre-training, providing BiTimeBERT with an unintended advantage. In contrast, BiTimeBERT 2.0 outperforms the original BiTimeBERT on the TDA-Timestamp dataset in most cases, particularly at month granularity, where it consistently achieves the best results. The TDA-Timestamp dataset poses a significant challenge, with 2,627 month labels spanning a wide temporal range from 1785 to 2009. These results highlight BiTimeBERT 2.0’s capability to effectively handle challenging datasets that extend far beyond the temporal scope of its pre-training corpus, consistent with its strong performance observed on WOTD and WOTD-WithCI datasets.

\begin{table}[]
\caption{Performance comparison of different models on the document dating task, evaluated using the NYT-Timestamp and TDA-Timestamp datasets. Bold font indicates the best performing model while underline is used to mark the second best one. Asterisk (*) denotes statistical significance at \(p < 0.01\), and dagger (†) denotes significance at \(p < 0.001\).}
\vspace{-1.0em}
\label{tab_main2}
\footnotesize
\renewcommand{\arraystretch}{1.3}
 \centering 
\begin{tabular}{|ccccccccc|}
\hline
\multicolumn{1}{|c|}{\multirow{3}{*}{\textbf{Model}}} & \multicolumn{4}{c|}{\textbf{NYT-Timestamp}} & \multicolumn{4}{c|}{\textbf{TDA-Timestamp}} \\ \cline{2-9} 
\multicolumn{1}{|c|}{} & \multicolumn{2}{c|}{\textbf{Year}} & \multicolumn{2}{c|}{\textbf{Month}} & \multicolumn{2}{c|}{\textbf{Year}} & \multicolumn{2}{c|}{\textbf{Month}} \\ \cline{2-9} 
\multicolumn{1}{|c|}{} & \multicolumn{1}{c|}{\textbf{ACC}} & \multicolumn{1}{c|}{\textbf{MAE}} & \multicolumn{1}{c|}{\textbf{ACC}} & \multicolumn{1}{c|}{\textbf{MAE}} & \multicolumn{1}{c|}{\textbf{ACC}} & \multicolumn{1}{c|}{\textbf{MAE}} & \multicolumn{1}{c|}{\textbf{ACC}} & \textbf{MAE} \\ \hline
\multicolumn{1}{|c|}{RG} & \multicolumn{1}{c|}{4.77} & \multicolumn{1}{c|}{7.06} & \multicolumn{1}{c|}{0.41} & \multicolumn{1}{c|}{81.79} & \multicolumn{1}{c|}{0.45} & \multicolumn{1}{c|}{75.39} & \multicolumn{1}{c|}{0.04} & 873.88 \\ \hline
\multicolumn{1}{|c|}{BERT-Orig} & \multicolumn{1}{c|}{35.00} & \multicolumn{1}{c|}{1.64} & \multicolumn{1}{c|}{2.56} & \multicolumn{1}{c|}{22.74} & \multicolumn{1}{c|}{15.84} & \multicolumn{1}{c|}{44.87} & \multicolumn{1}{c|}{0.80} & 632.66 \\ \hline
\multicolumn{9}{|c|}{\textit{Models Pre-trained using the entire NYT corpus}} \\ \hline
\multicolumn{1}{|c|}{BERT-NYT} & \multicolumn{1}{c|}{38.74} & \multicolumn{1}{c|}{1.41} & \multicolumn{1}{c|}{8.24} & \multicolumn{1}{c|}{18.35} & \multicolumn{1}{c|}{15.04} & \multicolumn{1}{c|}{45.16} & \multicolumn{1}{c|}{0.66} & 669.02 \\ \hline
\multicolumn{1}{|c|}{BERT+TIR} & \multicolumn{1}{c|}{48.06} & \multicolumn{1}{c|}{1.09} & \multicolumn{1}{c|}{20.30} & \multicolumn{1}{c|}{13.54} & \multicolumn{1}{c|}{17.72} & \multicolumn{1}{c|}{43.53} & \multicolumn{1}{c|}{1.26} & 589.69 \\ \hline
\multicolumn{1}{|c|}{BiTimeBERT} & \multicolumn{1}{c|}{\textbf{58.72}} & \multicolumn{1}{c|}{\textbf{0.80}} & \multicolumn{1}{c|}{\textbf{31.10}} & \multicolumn{1}{c|}{\textbf{9.54}} & \multicolumn{1}{c|}{{\ul 19.00}} & \multicolumn{1}{c|}{\textbf{40.11}} & \multicolumn{1}{c|}{{\ul 2.38}} & 580.25 \\ \hline
\multicolumn{9}{|c|}{\textit{Models Pre-trained using the RNYT corpus}} \\ \hline

\multicolumn{1}{|c|}{BERT-RNYT} & \multicolumn{1}{c|}{39.35} & \multicolumn{1}{c|}{1.48} & \multicolumn{1}{c|}{12.79} & \multicolumn{1}{c|}{19.40} & \multicolumn{1}{c|}{16.23} & \multicolumn{1}{c|}{44.53} & \multicolumn{1}{c|}{0.65} & 673.23 \\ \hline
\multicolumn{1}{|c|}{BiTimeBERT-RNYT} & \multicolumn{1}{c|}{50.12} & \multicolumn{1}{c|}{1.08} & \multicolumn{1}{c|}{23.48} & \multicolumn{1}{c|}{13.52} & \multicolumn{1}{c|}{16.58} & \multicolumn{1}{c|}{43.95} & \multicolumn{1}{c|}{1.07} & {\ul 571.09} \\ \hline
\multicolumn{1}{|c|}{BiTimeBERT+TSEMLM} & \multicolumn{1}{c|}{49.53} & \multicolumn{1}{c|}{1.15} & \multicolumn{1}{c|}{23.30} & \multicolumn{1}{c|}{13.25} & \multicolumn{1}{c|}{17.64} & \multicolumn{1}{c|}{42.75} & \multicolumn{1}{c|}{1.12} & 629.17 \\ \hline
\multicolumn{1}{|c|}{BiTimeBERT+TRWR} & \multicolumn{1}{c|}{51.63} & \multicolumn{1}{c|}{1.14} & \multicolumn{1}{c|}{25.35} & \multicolumn{1}{c|}{13.02} & \multicolumn{1}{c|}{17.48} & \multicolumn{1}{c|}{43.04} & \multicolumn{1}{c|}{1.36} & 575.40 \\ \hline
\multicolumn{1}{|c|}{BiTimeBERT 2.0} & \multicolumn{1}{c|}{{\ul 54.45}$^{\dagger}$} & \multicolumn{1}{c|}{{\ul 1.05}} & \multicolumn{1}{c|}{{\ul 29.07}$^{\dagger}$} & \multicolumn{1}{c|}{{\ul 10.42}$^{\dagger}$} & \multicolumn{1}{c|}{\textbf{19.92}$^{\dagger}$} & \multicolumn{1}{c|}{{\ul 42.14}$^{*}$} & \multicolumn{1}{c|}{\textbf{2.44}$^{\dagger}$} & \textbf{554.11}$^{*}$ \\ \hline
\end{tabular}

\end{table}

\subsubsection{Semantic Change Detection} 

\begin{table}[]
\caption{Performance comparison of different models on the semantic change detection task, evaluated using the LiverpoolFC and SemEval-Eng datasets. Bold font indicates the best performing model while underline is used to mark the second best one. Asterisk (*) denotes statistical significance at \(p < 0.01\), and dagger (†) denotes significance at \(p < 0.001\).}
\label{tab_main3}
\vspace{-1.0em}
\footnotesize
\renewcommand{\arraystretch}{1.3}
 \centering 
\begin{tabular}{|ccccc|}
\hline
\multicolumn{1}{|c|}{\multirow{2}{*}{\textbf{Model}}} & \multicolumn{2}{c|}{\textbf{LiverpoolFC}} & \multicolumn{2}{c|}{\textbf{SemEval-Eng}} \\ \cline{2-5} 
\multicolumn{1}{|c|}{} & \multicolumn{1}{c|}{\textbf{Pearson}} & \multicolumn{1}{c|}{\textbf{Spearman}} & \multicolumn{1}{c|}{\textbf{Pearson}} & \textbf{Spearman} \\ \hline
\multicolumn{1}{|c|}{BERT-Orig} & \multicolumn{1}{c|}{0.414} & \multicolumn{1}{c|}{0.454} & \multicolumn{1}{c|}{0.483} & 0.416 \\ \hline
\multicolumn{1}{|c|}{TempoBERT (cos\_dist)} & \multicolumn{1}{c|}{0.371} & \multicolumn{1}{c|}{0.451} & \multicolumn{1}{c|}{0.538} & 0.467 \\ \hline
\multicolumn{1}{|c|}{TempoBERT (time-diff)} & \multicolumn{1}{c|}{\textbf{0.637}} & \multicolumn{1}{c|}{\textbf{0.620}} & \multicolumn{1}{c|}{0.208} & 0.381 \\ \hline
\multicolumn{5}{|c|}{\textit{Models Pre-trained using the entire NYT corpus}} \\ \hline
\multicolumn{1}{|c|}{BERT-NYT} & \multicolumn{1}{c|}{0.431} & \multicolumn{1}{c|}{0.463} & \multicolumn{1}{c|}{0.510} & 0.422 \\ \hline
\multicolumn{1}{|c|}{BiTimeBERT} & \multicolumn{1}{c|}{0.468} & \multicolumn{1}{c|}{0.476} & \multicolumn{1}{c|}{{\ul 0.616}} & {\ul 0.476} \\ \hline
\multicolumn{5}{|c|}{\textit{Models Pre-trained using the RNYT corpus}} \\ \hline

\multicolumn{1}{|c|}{BERT-RNYT} & \multicolumn{1}{c|}{0.436} & \multicolumn{1}{c|}{0.451} & \multicolumn{1}{c|}{0.557} &  0.420\\ \hline
\multicolumn{1}{|c|}{BiTimeBERT-RNYT} & \multicolumn{1}{c|}{0.439} & \multicolumn{1}{c|}{0.460} & \multicolumn{1}{c|}{0.583} & 0.406 \\ \hline
\multicolumn{1}{|c|}{BiTimeBERT+TSETAMLM} & \multicolumn{1}{c|}{0.428} & \multicolumn{1}{c|}{0.424} & \multicolumn{1}{c|}{0.590} & 0.471 \\ \hline
\multicolumn{1}{|c|}{BiTimeBERT+TRWR} & \multicolumn{1}{c|}{0.474} & \multicolumn{1}{c|}{0.433} & \multicolumn{1}{c|}{0.602} & 0.421 \\ \hline
\multicolumn{1}{|c|}{BiTimeBERT 2.0} & \multicolumn{1}{c|}{{\ul 0.493}$^{\dagger}$} & \multicolumn{1}{c|}{{\ul 0.497}$^{\dagger}$} & \multicolumn{1}{c|}{\textbf{0.632}$^{\dagger}$} & \textbf{0.509}$^{\dagger}$ \\ \hline
\end{tabular}
\end{table}

Table~\ref{tab_main3} presents the results of the semantic change detection task on the \textbf{LiverpoolFC} and \textbf{SemEval-English} datasets, which encompass short-term and long-term corpora, respectively. For comparison, we include TempoBERT \cite{rosin2022time}, a time-aware BERT variant that preprocesses input texts by appending timestamp information and masking it during training, achieving state-of-the-art results in semantic change detection. Notably, while TempoBERT employs both cos\_dist (cosine distance) and a tailored time-diff (time difference) method for measuring semantic change, BiTimeBERT 2.0 and other models primarily rely on standard cos\_dist to compute Pearson and Spearman correlation scores. 
While TempoBERT achieves strong results on the short-term LiverpoolFC corpus using the time-diff method, it experiences a significant performance drop on the long-term SemEval-English corpus when applying the same method. This highlights the importance of selecting an appropriate measurement method (cos\_dist or time-diff) for TempoBERT, depending on the temporal span of the corpus (long-term or short-term), to ensure effective semantic change detection. In contrast, BiTimeBERT 2.0 demonstrates relatively stable performance and achieves the best results across both datasets when evaluated using the standard cos\_dist method. For example, compared to TempoBERT on the LiverpoolFC dataset, BiTimeBERT 2.0 improves Pearson correlation scores by 32.88\% and Spearman scores by 10.19\%. On the SemEval-English dataset, it achieves improvements of 17.47\% in Pearson and 8.99\% in Spearman scores. These gains can be attributed to BiTimeBERT 2.0's comprehensive pre-training approach, which integrates content temporal information, timestamp information, and time-sensitive entity data,  in contrast to TempoBERT’s exclusive focus on timestamp information. Furthermore, BiTimeBERT 2.0 surpasses the original BiTimeBERT, with improvements on SemEval-English of 2.59\% in Pearson and 6.93\% in Spearman scores, and on LiverpoolFC of 5.34\% in Pearson and 4.41\% in Spearman scores. These results underscore the effectiveness of the enhanced pre-training objectives in BiTimeBERT 2.0, particularly its ability to handle semantic changes across diverse temporal spans, further solidifying its robustness in time-aware language understanding.

\subsection{Additional Analysis} 
\subsubsection{Ablation Study} 
\label{Ablation}

\begin{table}[]
\caption{Ablation test on event occurrence time estimation task. All models are pre-trained on the RNYT corpus for 3 epochs. Bold font indicates the best performing model while underline is used to mark the second best one. Asterisk (*) denotes statistical significance at \(p < 0.01\), and dagger (†) denotes significance at \(p < 0.001\).}
\label{tab_abl1}

\renewcommand{\arraystretch}{1.3}
\vspace{-1.0em}
\footnotesize
\begin{tabular}{|c|cccc|cccc|cc|cc|}
\hline
\multirow{3}{*}{\textbf{Model}} & \multicolumn{4}{c|}{\textbf{EventTime}} & \multicolumn{4}{c|}{\textbf{EventTime-WithRelDoc}} & \multicolumn{2}{c|}{\textbf{WOTD}} & \multicolumn{2}{c|}{\textbf{WOTD-WithCI}} \\ \cline{2-13} 
 & \multicolumn{2}{c|}{\textbf{Year}} & \multicolumn{2}{c|}{\textbf{Month}} & \multicolumn{2}{c|}{\textbf{Year}} & \multicolumn{2}{c|}{\textbf{Month}} & \multicolumn{1}{c|}{\multirow{2}{*}{\textbf{ACC}}} & \multirow{2}{*}{\textbf{MAE}} & \multicolumn{1}{c|}{\multirow{2}{*}{\textbf{ACC}}} & \multirow{2}{*}{\textbf{MAE}} \\ \cline{2-9}
 & \multicolumn{1}{c|}{\textbf{ACC}} & \multicolumn{1}{c|}{\textbf{MAE}} & \multicolumn{1}{c|}{\textbf{ACC}} & \textbf{MAE} & \multicolumn{1}{c|}{\textbf{ACC}} & \multicolumn{1}{c|}{\textbf{MAE}} & \multicolumn{1}{c|}{\textbf{ACC}} & \textbf{MAE} & \multicolumn{1}{c|}{} &  & \multicolumn{1}{c|}{} &  \\ \hline
MLM & \multicolumn{1}{c|}{20.33} & \multicolumn{1}{c|}{3.65} & \multicolumn{1}{c|}{5.17} & 44.25 & \multicolumn{1}{c|}{37.15} & \multicolumn{1}{c|}{3.02} & \multicolumn{1}{c|}{11.28} & 32.63 & \multicolumn{1}{c|}{7.80} & 57.69 & \multicolumn{1}{c|}{16.37} & 34.65 \\ \hline
ETAMLM & \multicolumn{1}{c|}{22.41} & \multicolumn{1}{c|}{3.47} & \multicolumn{1}{c|}{6.70} & 43.12 & \multicolumn{1}{c|}{38.50}& \multicolumn{1}{c|}{2.75} & \multicolumn{1}{c|}{12.63} & 30.61 & \multicolumn{1}{c|}{8.92} & 54.03 & \multicolumn{1}{c|}{18.46} & 30.77 \\ \hline
DD & \multicolumn{1}{c|}{23.37} & \multicolumn{1}{c|}{3.72} & \multicolumn{1}{c|}{8.14} & 41.38 & \multicolumn{1}{c|}{39.52} & \multicolumn{1}{c|}{2.76} & \multicolumn{1}{c|}{15.42} & 31.69 & \multicolumn{1}{c|}{9.52} & 62.12 & \multicolumn{1}{c|}{17.41} & 43.22 \\ \hline
TSER & \multicolumn{1}{c|}{20.95} & \multicolumn{1}{c|}{3.66} & \multicolumn{1}{c|}{5.74} & 47.24 & \multicolumn{1}{c|}{37.36}& \multicolumn{1}{c|}{2.75} & \multicolumn{1}{c|}{9.25} & 32.17 & \multicolumn{1}{c|}{6.85} & 54.26 & \multicolumn{1}{c|}{14.93} & 43.05 \\ \hline

ETAMLM+DD & \multicolumn{1}{c|}{25.65} & \multicolumn{1}{c|}{{\ul 3.27}} & \multicolumn{1}{c|}{{\ul 8.48}} & {\ul 38.43} & \multicolumn{1}{c|}{39.10} & \multicolumn{1}{c|}{2.70} & \multicolumn{1}{c|}{{\ul 19.51}} & {\ul 30.07} & \multicolumn{1}{c|}{10.31} & 55.50 & \multicolumn{1}{c|}{ 22.08} & \textbf{33.01} \\ \hline
ETAMLM+TSER & \multicolumn{1}{c|}{23.83} & \multicolumn{1}{c|}{3.51} & \multicolumn{1}{c|}{7.23} & 42.20 & \multicolumn{1}{c|}{{\ul 40.30}} & \multicolumn{1}{c|}{{\ul 2.69}} & \multicolumn{1}{c|}{16.20} & 30.15 & \multicolumn{1}{c|}{10.14} & {\ul 53.91} & \multicolumn{1}{c|}{22.06} & 33.42 \\ \hline
DD+TSER & \multicolumn{1}{c|}{{\ul 25.78}} & \multicolumn{1}{c|}{3.40} & \multicolumn{1}{c|}{8.33} & 39.61 & \multicolumn{1}{c|}{40.12} & \multicolumn{1}{c|}{\textbf{2.61}} & \multicolumn{1}{c|}{18.06} & 31.40 & \multicolumn{1}{c|}{{\ul 11.10}} & 57.89 & \multicolumn{1}{c|}{{\ul 22.42}} & 39.65 \\ \hline
ETAMLM+DD+TSER & \multicolumn{1}{c|}{\textbf{27.24}$^{\dagger}$} & \multicolumn{1}{c|}{\textbf{3.18}$^*$} & \multicolumn{1}{c|}{\textbf{10.11}$^{\dagger}$} & \textbf{38.34} & \multicolumn{1}{c|}{\textbf{42.04}$^{\dagger}$} & \multicolumn{1}{c|}{2.71} & \multicolumn{1}{c|}{\textbf{19.92}$^{*}$} & \textbf{29.64} & \multicolumn{1}{c|}{\textbf{11.97}$^{\dagger}$} & \textbf{50.23}$^{\dagger}$ & \multicolumn{1}{c|}{\textbf{23.59}$^{\dagger}$} & {\ul 33.24} \\ \hline
\end{tabular}

\vspace{1.5em}

\caption{Ablation test on document dating task. All models are pre-trained on the RNYT corpus for 3 epochs. Bold font indicates the best performing model while underline is used to mark the second best one. Asterisk (*) denotes statistical significance at \(p < 0.01\), and dagger (†) denotes significance at \(p < 0.001\).}
\label{tab_abl2}
\renewcommand{\arraystretch}{1.3}
\vspace{-1.0em}
\footnotesize
\begin{tabular}{|c|cccc|cc|}
\hline
\multirow{3}{*}{\textbf{Model}} & \multicolumn{4}{c|}{\textbf{NYT-Timestamp}} & \multicolumn{2}{c|}{\textbf{TDA-Timestamp}} \\ \cline{2-7} 
 & \multicolumn{2}{c|}{\textbf{Year}} & \multicolumn{2}{c|}{\textbf{Month}} & \multicolumn{2}{c|}{\textbf{Year}} \\ \cline{2-7} 
 & \multicolumn{1}{c|}{\textbf{ACC}} & \multicolumn{1}{c|}{\textbf{MAE}} & \multicolumn{1}{c|}{\textbf{ACC}} & \textbf{MAE} & \multicolumn{1}{c|}{\textbf{ACC}} & \textbf{MAE} \\ \hline

MLM & \multicolumn{1}{c|}{32.58} & \multicolumn{1}{c|}{1.50} & \multicolumn{1}{c|}{6.07} & 18.37 & \multicolumn{1}{c|}{13.03} & 43.62 \\ \hline
ETAMLM & \multicolumn{1}{c|}{34.71} & \multicolumn{1}{c|}{1.32} & \multicolumn{1}{c|}{10.22} & 17.36 & \multicolumn{1}{c|}{15.04} & 42.51 \\ \hline
DD & \multicolumn{1}{c|}{45.25} & \multicolumn{1}{c|}{1.46} & \multicolumn{1}{c|}{19.77} & 15.37 & \multicolumn{1}{c|}{15.63} & 45.02 \\ \hline
TSER & \multicolumn{1}{c|}{32.94} & \multicolumn{1}{c|}{1.42} & \multicolumn{1}{c|}{8.58} & 19.42 & \multicolumn{1}{c|}{15.24} & 44.72 \\ \hline

ETAMLM+DD & \multicolumn{1}{c|}{{\ul 48.08}} & \multicolumn{1}{c|}{1.17} & \multicolumn{1}{c|}{21.29} & {\ul 13.93} & \multicolumn{1}{c|}{16.12} & 44.80 \\ \hline
ETAMLM+TSER & \multicolumn{1}{c|}{40.12} & \multicolumn{1}{c|}{1.32} & \multicolumn{1}{c|}{19.44} & 14.26 & \multicolumn{1}{c|}{16.85} & {\ul 41.37} \\ \hline
DD+TSER & \multicolumn{1}{c|}{47.90} & \multicolumn{1}{c|}{\textbf{1.15}} & \multicolumn{1}{c|}{{\ul 22.39}} & 14.05 & \multicolumn{1}{c|}{{\ul 17.62}} & \textbf{41.25} \\ \hline
ETAMLM+DD+TSER & \multicolumn{1}{c|}{\textbf{50.00}$^{\dagger}$} & \multicolumn{1}{c|}{{\ul 1.16}}  & \multicolumn{1}{c|}{\textbf{24.46}$^{\dagger}$} & \textbf{13.31}$^{*}$ & \multicolumn{1}{c|}{\textbf{18.29}$^{\dagger}$} & 42.33$^{\dagger}$ \\ \hline
\end{tabular}
\end{table}

To evaluate the contributions of the three pre-training objectives in BiTimeBERT 2.0, we conduct an ablation study, with the results presented in Table~\ref{tab_abl1} (event occurrence time estimation) and Table~\ref{tab_abl2} (document dating). This analysis compares eight distinct model configurations, each employing different combinations of pre-training tasks, tested across various datasets to evaluate their effectiveness. The configurations include models trained with individual objectives (\textit{MLM}, \textit{ETAMLM}, \textit{DD}, or \textit{TSER}) as well as combinations of two or all three objectives. 
For instance, the \textit{ETAMLM+DD} model combines ETAMLM and DD objectives, while \textit{ETAMLM+TSER} and \textit{DD+TSER} pair ETAMLM with TSER and DD with TSER, respectively. The complete BiTimeBERT 2.0 model, represented as \textit{ETAMLM+DD+TSER}, incorporates all three proposed objectives. To ensure a fair and effective comparison, all model variants are continually pre-trained on the RNYT corpus for three epochs, using the $BERT_{BASE}$ (cased) model as the starting point. Notably, the TDA-Timestamp dataset at month granularity is excluded due to consistently poor performance across all models. 

When examining models utilizing a single pre-training objective, the ETAMLM model consistently outperforms the MLM model, demonstrating the enhanced effectiveness of leveraging time-aware masking strategies to capture temporal nuances in the data. Additionally, the DD model generally surpasses other single-objective models, particularly excelling in the document dating task. This result highlights the advantage of integrating task-specific knowledge during pre-training, effectively aligning pre-training with downstream tasks. Although TSER as a standalone objective is less effective compared to ETAMLM and DD, combining it with these objectives, such as in ETAMLM+TSER or DD+TSER, significantly enhances performance. This indicates the value of incorporating time-sensitive entity information for a more holistic understanding of temporal dynamics. For instance, on the WOTD-WithCI dataset, the DD+TSER model achieves improvements of 50.16\% in ACC and 7.89\% in MAE compared to TSER alone, and gains of 28.77\% in ACC and 8.26\% in MAE when compared to DD alone. These observations are consistent across other cases, where models incorporating two pre-training objectives generally outperform those trained with a single objective. Notably, the ETAMLM+DD model consistently ranks second-best in most scenarios. However, the ETAMLM+DD+TSER model, incorporating all three pre-training tasks of BiTimeBERT 2.0, achieves the best results across the four datasets in most cases. This underscores the collective contribution of ETAMLM, DD, and TSER to the model's performance, enabling the generation of highly effective time-aware language representations and confirming the value of integrating diverse temporal dimensions into pre-training.

\subsubsection{Analysis of Event Type Distributions, Prediction Accuracy, and Recall Ratios in Time Estimation Tasks.} 

\begin{figure}
    \centering
    \includegraphics[width=0.90\linewidth]{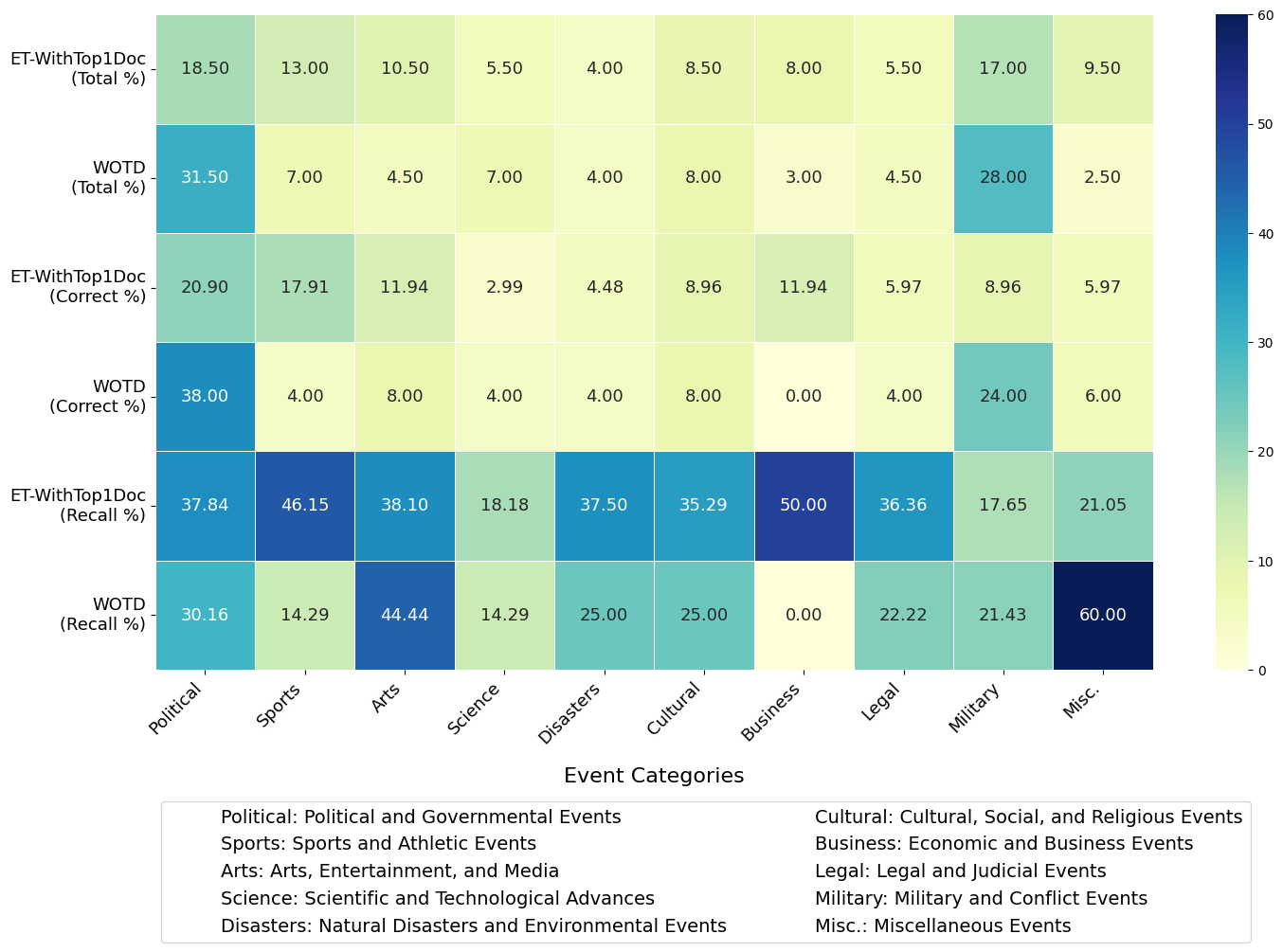}
    \caption{Comparative Heatmap of Event Type Distributions, Prediction Accuracy, and Recall Ratios.}
    \label{EventCate}
\end{figure}

In this analysis, we investigate the types of events that BiTimeBERT 2.0 accurately predicts, with a particular focus on the EventTime-WithRelDoc and WOTD datasets at the year-level granularity.
From each dataset, we randomly select 200 instances and manually categorize these events into 10 distinct types: "Political and Governmental Events", "Sports and Athletic Events", "Arts, Entertainment, and Media", "Scientific and Technological Advances", "Natural Disasters and Environmental Events", "Cultural, Social, and Religious Events", "Economic and Business Events", "Legal and Judicial Events", "Military and Conflict Events", and "Miscellaneous Events". The heatmap in Figure~\ref{EventCate} offers a comprehensive visualization of event type distributions, prediction accuracy, and recall ratios across these categories. 
The first two rows of the heatmap represent the percentage distributions of event types (Total \%) in the EventTime-WithRelDoc and WOTD datasets, showing how the events are distributed within the sampled data. The next two rows depict the percentage of correctly predicted events for each category (Correct \%), providing insights into the model’s prediction performance. Finally, the last two rows display the recall ratios (Recall \%), which measure the proportion of correctly predicted events relative to the total number of instances in each category. For example, in the EventTime-WithRelDoc dataset, the 200 sampled instances include 37 "Political and Governmental Events" instances, which account for 18.50\% (37/200) of the total distribution. BiTimeBERT 2.0 accurately predicts 67 events overall in this dataset, including 14 correctly predicted instances within the "Political and Governmental Events" category. This results in a prediction accuracy of 20.90\% (14/67) for this category and a recall ratio of 37.83\% (14/37).

When analyzing the first two rows of the heatmap, which depict the distributions of event types (Total \%), we observe that the EventTime-WithRelDoc dataset is more evenly distributed, with four categories each exceeding 10\%. 
In contrast, the WOTD dataset exhibits a more skewed distribution. 
However, in both datasets, "Political and Governmental Events" and "Military and Conflict Events" are the two most prevalent categories. 
When examining the correctly predicted events (Correct \%), the heatmap shows that "Political and Governmental Events" maintain their prominence across both datasets. This category consistently accounts for a significant proportion of the correctly predicted instances, reflecting the model’s strength in handling this type of event. For example, in the EventTime-WithRelDoc dataset, "Political and Governmental Events" contribute 20.90\% of the correctly predicted events, closely aligning with their original distribution in the dataset. However, the percentage of correctly predicted events for "Military and Conflict Events" is noticeably lower than their share in the total distribution, particularly in the EventTime-WithRelDoc dataset. This discrepancy suggests that BiTimeBERT 2.0 struggles to accurately predict events in this category, potentially due to the inherent complexity or ambiguity associated with their temporal patterns.
The recall ratios (Recall \%), depicted in the final two rows of the heatmap, provide additional insight into the model’s predictive capabilities across event categories. In the EventTime-WithRelDoc dataset, recall ratios are relatively consistent across most event types, indicating that the model performs robustly within the temporal scope of its pre-training corpus. 
In contrast, the WOTD dataset, which spans a broader historical range and includes more temporally diverse events, demonstrates greater variability in recall ratios. While categories like "Political and Governmental Events" and "Arts, Entertainment, and Media" achieve relatively high recall values, others such as "Sports and Athletic Events" show notably lower recall. 
This variability underscores that while BiTimeBERT 2.0 outperforms other models on datasets extending beyond its pre-training temporal scope, it still struggles with certain event types. To address these limitations, further training on underperforming categories, combined with strategies such as incremental training and task-specific refinements (which will be detailed in Section \ref{section5_2_3} ), could improve the model’s ability to generalize across broader temporal contexts and challenging event types.

\begin{figure}[h]
    \centering
    \begin{minipage}{0.49\textwidth}
        \centering
        \includegraphics[width=\textwidth]{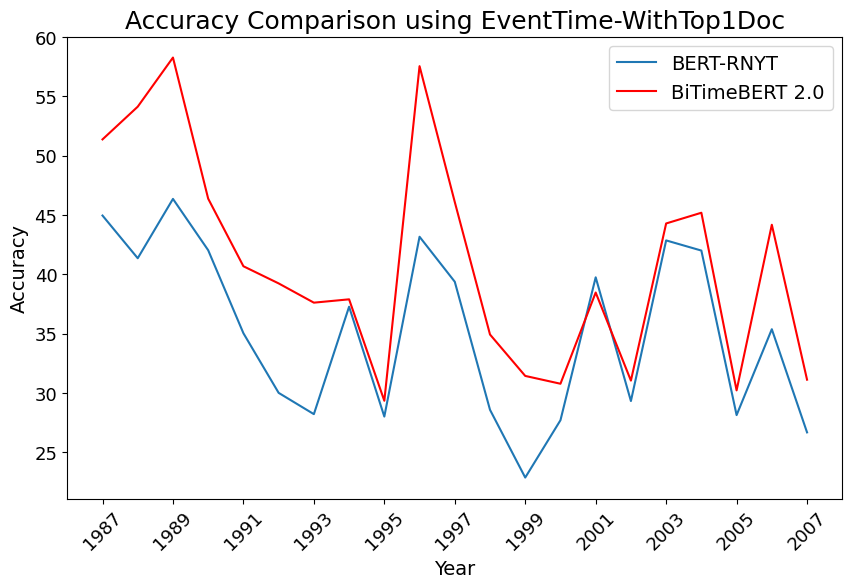} 
\vspace{-1.0em}
        \label{AC1}
    \end{minipage}\hfill
    \begin{minipage}{0.49\textwidth}
        \centering
        \includegraphics[width=\textwidth]{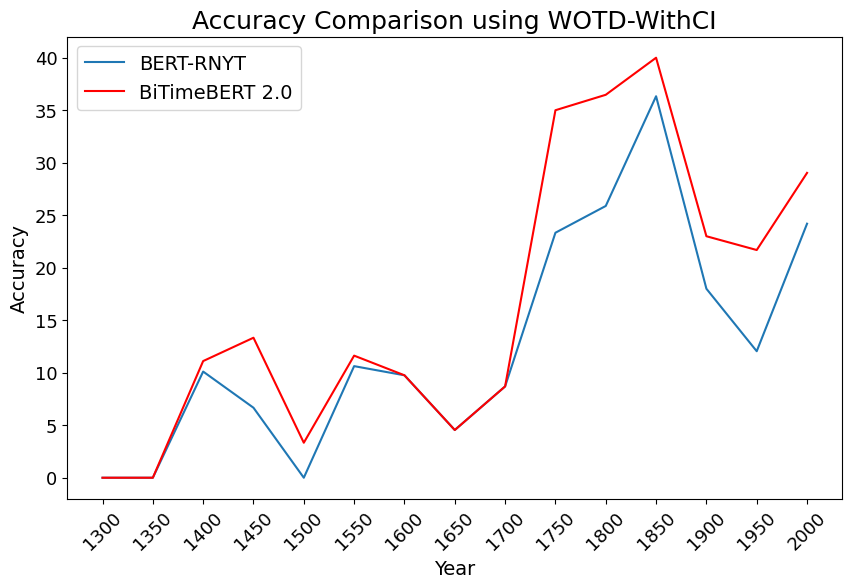} 
\vspace{-1.5em}
    \end{minipage}

\caption{Accuracy Trends of BiTimeBERT 2.0 and BERT-RNYT Over Time}
        \label{figure:AC}
\end{figure}

\subsubsection{Model Accuracy Trends Over Time.} \label{section5_2_3} This section explores the accuracy trends of BiTimeBERT 2.0 and BERT-RNYT in estimating event occurrence times, based on the main results presented in Section \ref{section5_1_1}. Figure~\ref{figure:AC} plots the accuracy trends for both models on the EventTime-WithRelDoc and WOTD-WithCI datasets under year granularity, which exhibit distinct temporal characteristics. EventTime-WithRelDoc aligns with the temporal range of the pre-training NYT corpus (1987–2007), whereas WOTD-WithCI spans a much broader period, from 1302 to 2018. For EventTime-WithRelDoc, we analyze yearly accuracy from 1987 to 2007, while for WOTD-WithCI, accuracy is evaluated in 50-year intervals due to its extensive temporal span. Our findings reveal that BiTimeBERT 2.0 generally outperforms BERT-RNYT across both datasets over time, underscoring the effectiveness of BiTimeBERT 2.0's three integrated objectives. On EventTime-WithRelDoc, BiTimeBERT 2.0 exhibits notably higher accuracy in the earlier years, particularly in 1989 and 1997, with a smaller margin of improvement in later years. On WOTD-WithCI, BiTimeBERT 2.0 either matches or surpasses BERT-RNYT at every evaluated time point. The largest improvements occur between 1700 and 1850 and again from 1900 onward, with many of these periods falling outside the temporal scope of the pre-training corpus. These results underscore BiTimeBERT 2.0's strong ability to generalize to both in-range and out-of-range temporal contexts.

These results, along with BiTimeBERT 2.0's strong performance on datasets spanning vastly different temporal ranges (e.g., WOTD, WOTD-WithCI, TDA-Timestamp, LiverpoolFC, and SemEval-English), highlight its enhanced generalization capabilities. 
However, while the model performs well on datasets with broad temporal spans, its adaptability to more recent data remains a critical area for improvement, particularly for contemporary and future temporal contexts. Incremental training on recent datasets presents a promising avenue to address this limitation. This process involves periodically updating the training corpus with newly published articles and datasets that reflect current events, emerging entities, and evolving temporal trends. By continually integrating modern temporal signals, the model can stay relevant to recent data without being constrained by its original pre-training timeframe.
Task-specific refinements to BiTimeBERT 2.0’s pre-training objectives could further enhance its adaptability. For instance, the ETAMLM task could prioritize masking tokens corresponding to modern temporal spans, ensuring the model effectively learns recent trends and patterns. Similarly, the DD task could refine its prediction accuracy by aligning it with timestamps from newly introduced data, and the TSER objective could incorporate dynamic entities from contemporary contexts to maintain relevance. 
A key challenge in this approach is avoiding catastrophic forgetting of earlier temporal knowledge when adapting the model to recent data. Addressing this requires balancing the retention of older knowledge with the integration of new information. Techniques such as experience replay \cite{rolnick2019experience} or regularization-based techniques like Elastic Weight Consolidation (EWC) \cite{kirkpatrick2017overcoming, lee2017overcoming} can be employed to mitigate the risk of losing earlier knowledge during incremental updates. These approaches help the model retain its strong performance on older temporal ranges while adapting to new data.
Future research could focus on integrating these strategies with continuous validation against temporally diverse datasets. Furthermore, while adapting to recent data is essential, ensuring robust performance on older datasets, including those from periods far earlier than 1987, is equally important. Similar strategies, such as incremental training and task-specific refinements, could be extended to historical data. For instance, the ETAMLM objective could emphasize masking tokens associated with older temporal spans, while the TSER objective could integrate historical entities and contexts to better align with past events. Incremental updates incorporating historical corpora would further enhance the model's ability to generalize across a wide temporal spectrum. 
By balancing the integration of recent and historical temporal knowledge, BiTimeBERT 2.0 could not only evolve with contemporary trends but also excel in handling datasets spanning vast historical periods.

\refstepcounter{subsubsection} 
\subsubsection*{\textbf{\thesubsubsection.\hspace{0.5em}Temporal Semantic Similarity Analysis}}
\label{section5_2_4}
\label{similarity_section}

In this analysis, we conduct straightforward similarity experiments to assess whether BiTimeBERT 2.0 generates effective time-aware language representations without fine-tuning for specific downstream tasks. To facilitate this evaluation, we utilize two datasets: the EventTime test set, which consists of 2,240 instances covering the temporal range from 1987 to 2007, and a subset of the WOTD test set, which encompasses 213 instances of events spanning from 1919 to 2018. 
The temporal semantic similarity analysis is performed at a yearly granularity. We begin by collecting contextual representations, specifically the final hidden state vector of the [CLS] token output by the model for all relevant atomic time units defined by each dataset. The resulting vector collection for the WOTD subset contains 100 vectors corresponding to yearly temporal expressions from 1919 to 2018, while the EventTime dataset yields 21 vectors for expressions ranging from 1987 to 2007. Following this step, we compute cosine similarities between the contextual representations of each event description and each yearly temporal expression. The temporal expression with the highest cosine similarity score is designated as the estimated event time for each event description.
To enhance our analysis, we further categorize the results from the WOTD dataset into three distinct time ranges: events occurring before the pre-training corpus ("1919-1986"), events within the pre-training corpus ("1987-2007"), and events occurring after the pre-training corpus ("2008-2018"). These time ranges encompass 128 instances for the first category, 58 for the second, and 27 for the third.

\begin{table}[]
\footnotesize
\caption{Temporal Semantic Similarity Analysis Results on EventTime and WOTD Dataset. Models are tested without fine-tuning.}
\renewcommand{\arraystretch}{1.1}
\begin{tabular}{|c|cc|cccccccc|}
\hline
\multirow{3}{*}{\textbf{Model}} & \multicolumn{2}{c|}{\textbf{EventTime}}             & \multicolumn{8}{c|}{\textbf{WOTD}}                                                                                                                                                                                                                                                   \\ \cline{2-11} 
                                & \multicolumn{2}{c|}{\textbf{1987-2007}}                      & \multicolumn{2}{c|}{\textbf{1919-2018}}                                  & \multicolumn{2}{c|}{\textbf{1919-1986}}                                  & \multicolumn{2}{c|}{\textbf{1987-2007}}                                  & \multicolumn{2}{c|}{\textbf{2008-2018}}             \\ \cline{2-11} 
                                & \multicolumn{1}{c|}{\textbf{ACC}}   & \textbf{MAE}  & \multicolumn{1}{c|}{\textbf{ACC}}  & \multicolumn{1}{c|}{\textbf{MAE}}   & \multicolumn{1}{c|}{\textbf{ACC}}  & \multicolumn{1}{c|}{\textbf{MAE}}   & \multicolumn{1}{c|}{\textbf{ACC}}   & \multicolumn{1}{c|}{\textbf{MAE}}  & \multicolumn{1}{c|}{\textbf{ACC}}   & \textbf{MAE}  \\ \hline
BERT                            & \multicolumn{1}{c|}{3.03}           & 10.47         & \multicolumn{1}{c|}{2.34}          & \multicolumn{1}{c|}{27.89}          & \multicolumn{1}{c|}{3.12}          & \multicolumn{1}{c|}{{\ul 23.10} }         & \multicolumn{1}{c|}{0.0}            & \multicolumn{1}{c|}{35.60}         & \multicolumn{1}{c|}{3.70}           & 34.0          \\ \hline
BERT-RNYT                       & \multicolumn{1}{c|}{5.08}           & 7.17          & \multicolumn{1}{c|}{2.73}          & \multicolumn{1}{c|}{28.76}          & \multicolumn{1}{c|}{3.78}          & \multicolumn{1}{c|}{34.90}          & \multicolumn{1}{c|}{1.72}           & \multicolumn{1}{c|}{17.29}         & \multicolumn{1}{c|}{0.0}            & 24.33         \\ \hline
BiTimeBERT                      & \multicolumn{1}{c|}{\textbf{14.33}} & {\ul 5.72}          & \multicolumn{1}{c|}{6.10}          & \multicolumn{1}{c|}{25.03}          & \multicolumn{1}{c|}{2.34}          & \multicolumn{1}{c|}{27.64}          & \multicolumn{1}{c|}{\textbf{17.24}} & \multicolumn{1}{c|}{17.77}         & \multicolumn{1}{c|}{0.0}            & 28.25         \\ \hline
BiTimeBERT-RNYT                 & \multicolumn{1}{c|}{10.62}          & 6.47          & \multicolumn{1}{c|}{{\ul 7.51}}          & \multicolumn{1}{c|}{{\ul 23.53}}          & \multicolumn{1}{c|}{{\ul 3.90}}          & \multicolumn{1}{c|}{32.16}          & \multicolumn{1}{c|}{13.79}          & \multicolumn{1}{c|}{\textbf{8.65}} & \multicolumn{1}{c|}{{\ul 7.40}}           & {\ul 14.59}         \\ \hline
BiTimeBERT 2.0                  & \multicolumn{1}{c|}{{\ul 13.08}}          & \textbf{5.46} & \multicolumn{1}{c|}{\textbf{9.85}} & \multicolumn{1}{c|}{\textbf{16.89}} & \multicolumn{1}{c|}{\textbf{4.68}} & \multicolumn{1}{c|}{\textbf{21.85}} & \multicolumn{1}{c|}{\textbf{17.24}} & \multicolumn{1}{c|}{{\ul 9.91}}          & \multicolumn{1}{c|}{\textbf{18.51}} & \textbf{8.40} \\ \hline
\end{tabular}
\label{tab:temporal_similarity_results}
\end{table}

The results of our temporal semantic similarity analysis are summarized in Table \ref{tab:temporal_similarity_results}, comparing the performance of various models on both the EventTime and WOTD datasets. For the EventTime dataset, BiTimeBERT 2.0 achieves the lowest MAE of 5.46, and slightly underperforms BiTimeBERT in terms of ACC. Conversely, on the WOTD dataset, BiTimeBERT 2.0 shows a significant advantage compared to BiTimeBERT, achieving a remarkable 61.47\% increase in ACC and a 32.52\% improvement in MAE across the 1919-2018 time range.
Additionally, BiTimeBERT 2.0 demonstrates even greater gains in time ranges beyond the pre-training corpus. For instance, in the 1919-1986 period, it achieves a notable 100\% improvement in ACC and a 20.94\% enhancement in MAE compared to BiTimeBERT. In the 2008-2018 range, BiTimeBERT 2.0 records an ACC of 18.51 and a MAE of 8.40, in stark contrast to BiTimeBERT's ACC of 0.0 and MAE of 24.33. BiTimeBERT-RNYT also outperforms BiTimeBERT in these ranges, obtaining the second-best results for the overall 1919-2018 analysis.
These findings affirm BiTimeBERT 2.0's capacity to effectively capture temporal nuances and construct robust time-aware language representations, exhibiting proficiency in learning domain-specific and task-oriented knowledge without the need for fine-tuning. Furthermore, its strong performance outside the pre-training corpus range highlights its robustness and adaptability.

\refstepcounter{subsection} 
\subsection*{\textbf{\thesubsection.\hspace{0.5em}Case Study on Time-Sensitive Queries}}

Following our previous work \cite{wang2023bitimebert}, we conduct two case studies focused on event time prediction, utilizing concise, time-sensitive queries related to specific events.\footnote{Notably, the average number of tokens of the queries here is only 3.2, in contrast to 17.3 tokens for the EventTime dataset and 17.6 tokens for the WOTD dataset employed in Section \ref{similarity_section}.} The queries encompass both non-recurring and recurring events. A non-recurring event refers to an event that occurred at a specific point in time (e.g., "9/11 attacks"), whereas a recurring event refers to those that take place multiple times in history (e.g., "Summer Olympic Games").
Similar to the methodology outlined in Section \ref{similarity_section}, we utilize the cosine similarity between the representation of each query and the corresponding temporal expressions, applying the model without fine-tuning. Instead of calculating ACC and MAE based solely on the date with the highest similarity score, we generate ranked lists of potential dates. For non-recurring event analysis, we compute the Mean Reciprocal Rank (MRR) to identify the position of the correct date within the list. In contrast, for recurring event analysis, we employ Mean Average Precision (MAP) to evaluate whether all occurrence dates of the recurring event are ranked near the top.

\label{section5_3}

\begin{table}
\footnotesize
\caption{Results for non-recurring (left) and recurring events (right). Models are tested without fine-tuning.}
\renewcommand{\arraystretch}{1.1}
\vspace{-0.2em}
\label{tab_casestudy}
\begin{tabular}{|c|c|cc|}
\hline
\multirow{3}{*}{\textbf{Model}} & \textbf{Non-Recurring Events} & \multicolumn{2}{c|}{\textbf{Recurring Events}}               \\ \cline{2-4} 
                                & \multirow{2}{*}{\textbf{MRR}}          & \multicolumn{1}{c|}{\textbf{1966-1986}} & \textbf{1987-2007} \\ \cline{3-4} 
                                &                               & \multicolumn{1}{c|}{\textbf{MAP}}       & \textbf{MAP}       \\ \hline
BERT                            & 0.1277                        & \multicolumn{1}{c|}{0.4042}             & 0.3512             \\ \hline
BERT-RNYT                       & 0.3733                        & \multicolumn{1}{c|}{0.4980}             & 0.4071             \\ \hline
BiTimeBERT                      & \textbf{0.5416}               & \multicolumn{1}{c|}{{\ul 0.5294}}       & {\ul 0.6686}      \\ \hline
BiTimeBERT-RNYT                 & 0.3737                        & \multicolumn{1}{c|}{0.4818}             & 0.5591             \\ \hline
BiTimeBERT 2.0                  & {\ul 0.4259}                & \multicolumn{1}{c|}{\textbf{0.5656}}    & \textbf{0.7176}    \\ \hline
\end{tabular}
\end{table}

\refstepcounter{subsubsection} 
\subsubsection*{\textbf{\thesubsubsection.\hspace{0.5em}Non-Recurring Events}}
For the non-recurring event analysis, we prepare ten short queries related to the September 11 attacks of 2001: "9/11 attacks", "Aircraft hijackings", "19 terrorists", "Osama bin Laden", "the Twin Towers", "War on terrorism", "American Airlines Flight 77", "American Airlines Flight 11", "United Airlines Flight 175", "United Airlines Flight 93". A ranked list is generated by comparing the cosine similarity between the query vector and the temporal expression vectors spanning the years from 1987 to 2007. As presented in Table~\ref{tab_casestudy}, BiTimeBERT achieved the highest MRR, with BiTimeBERT 2.0 closely following in second place.

\refstepcounter{subsubsection} 
\subsubsection*{\textbf{\thesubsubsection.\hspace{0.5em}Recurring Events}}
For the recurring event analysis, we collect ten short queries representing significant periodic events: "Summer Olympic Games", "FIFA World Cup", "Asian Games", "Commonwealth Games", "World Chess Championship", "United States presidential election", "French presidential election", "United Kingdom general election", "United States senate election", "United States midterm election". Given that these events occurred prior to 1987, we also compare them against temporal expressions spanning from 1966 to 1986, resulting in the creation of two distinct ranked lists of dates. As illustrated in Table~\ref{tab_casestudy}, BiTimeBERT 2.0 exhibits superior performance across both time periods, indicating that most occurrence dates are positioned at the top of the ranked lists. Additionally, when estimating dates within the temporal scope of the pre-training corpus (i.e., from 1987 to 2007), BiTimeBERT 2.0 also exhibits improved performance compared to BiTimeBERT.

These findings also suggest that BiTimeBERT 2.0 effectively integrates both domain knowledge and task-oriented information derived from temporal news during its pre-training phase. Consequently, the model excels at constructing effective representations that accurately capture the temporal aspects of queries, even when they are notably brief. This capability makes BiTimeBERT 2.0 a powerful tool for applications that require detailed temporal analysis.

\refstepcounter{subsection} 
\subsection*{\textbf{\thesubsection.\hspace{0.5em}Leveraging BiTimeBERT 2.0 for Enhanced Temporal Understanding}}
\refstepcounter{subsubsection} 
\subsubsection*{\textbf{\thesubsubsection.\hspace{0.5em}Temporal Question Answering with BiTimeBERT 2.0}}
BiTimeBERT 2.0 can be used in several ways and supports different applications for which time is important. In this section, we illustrate its integration into a temporal question answering system called QANA \cite{wang2021improving}, which demonstrates strong performance in addressing event-related questions that are implicitly time-scoped. For instance, the question, "Which famous painting by Norwegian Edvard Munch was stolen from the National Gallery in Oslo?" is inherently linked to a specific temporal event that occurred in May 1994. To effectively answer such implicitly time-scoped inquiries, QANA first estimates the time scope of the event described in the question at a monthly granularity, mapping it to a corresponding time interval defined by "start" and "end" dates (e.g., a potential time scope for this example is ("1994/03", "1994/08")). In the original implementation of QANA, the time scope estimation relies on analyzing the temporal distribution of the retrieved documents. We enhance this framework by incorporating a fine-tuned BiTimeBERT 2.0, specifically fine-tuned on the EventTime-WithRelDoc dataset at the monthly level. Following the methodology employed in constructing EventTime-WithRelDoc, we first identify the top relevant document for each question using the BM25 algorithm. The timestamp and content of this document are then appended to the corresponding question and processed as input for BiTimeBERT 2.0. The model subsequently predicts two time points with the highest probabilities, which are designated as the "start" and "end" markers for the estimated time scope of the question. This estimated time scope is then utilized to re-rank the documents, allowing the Document Reader Module of QANA to retrieve the final answers. Thus, our adaptation of QANA primarily modifies the time scope estimation step, and we refer to this revised version as QANA+BiTimeBERT 2.0.

\begin{table}
\begin{center}
\footnotesize
  \centering \caption{Performance of different models in QA task.}
  \vspace{-1.8em}
  \label{tab_QANA}
  \renewcommand{\arraystretch}{1.0}
  \setlength{\tabcolsep}{0.6em}
  \centering
  \begin{tabular}{|c|P{0.46cm}|P{0.46cm}|P{0.46cm}|P{0.46cm}|P{0.46cm}|P{0.46cm}|P{0.46cm}|P{0.46cm}|}
    \hline
    \multirow{2}{0.70cm}{\textbf{Model}} & \multicolumn{2}{c|}{\textbf{Top 1}} & \multicolumn{2}{c|}{\textbf{Top 5}}& \multicolumn{2}{c|}{\textbf{Top 10}}& \multicolumn{2}{c|}{\textbf{Top 15}}\\
    \cline{2-9}
    & \textbf{EM} & \textbf{F1} & \textbf{EM} & \textbf{F1} & \textbf{EM} & \textbf{F1} & \textbf{EM} & \textbf{F1}\\
    \hline
    QANA & 21.00& 28.90 & 28.20 & 36.85 & 34.20 & 44.01 & 36.20 & 45.63\\ \hline
    QANA + BiTimeBERT &  \textbf{22.40} & 29.31 & \textbf{29.20} & \textbf{37.14} & 34.80 & \textbf{44.34} & 36.40 & 46.01  \\ \hline
    QANA + BiTimeBERT 2.0 & 21.60 & \textbf{29.45} & 28.80 & 36.50 & \textbf{35.20} & 44.10 & \textbf{36.60} & \textbf{47.14} \\ \hline

  \end{tabular}
  \end{center}
\vspace{-1.5em} 
\end{table}

To evaluate the performance of this enhanced system, we test it on a dataset of 500 manually crafted implicitly time-scoped questions \cite{wang2021improving}. We examine the effects of varying the number of re-ranked documents by experimenting with different top \( N \) values. As indicated in Table~\ref{tab_QANA}, both QANA+BiTimeBERT 2.0 and QANA+BiTimeBERT consistently outperform the original QANA across all tested \( N \) values, highlighting their effectiveness. Both models facilitate temporal exploration of textual archives by estimating the temporal relevance of queries, enabling time-aware document ranking. Additionally, in Section \ref{section5_2_4} and Section \ref{section5_3}, we illustrate BiTimeBERT 2.0's remarkable success in estimating the temporal intent behind short event descriptions and queries, even without fine-tuning. In addition to its applications in temporal question answering, BiTimeBERT 2.0 has potential uses in various domains, including temporal information retrieval \cite{alonso2007value, campos2015survey}, event detection and ordering \cite{strotgen2012event, bookdercz}, document dating \cite{jatowt2012large, kanhabua2009using, kotsakos2014burstiness}, etc. In the following section, we discuss the potential application of BiTimeBERT 2.0 in temporal web analysis.

\refstepcounter{subsubsection} 
\subsubsection*{\textbf{\thesubsubsection.\hspace{0.5em}Exploring the Potential of BiTimeBERT 2.0 in Temporal Web Analysis}}
The rapid growth of digitally born content and extensive web archives offers valuable opportunities for temporal web analysis. BiTimeBERT 2.0, with its established effectiveness in time-sensitive tasks such as event occurrence time estimation, document dating, and semantic change detection, can be adapted to enhance the precision and relevance of temporal understanding within the temporal web domain. For instance, BiTimeBERT 2.0 can be utilized to improve temporal web search by aligning search results more closely with user intent. By inferring the implicit temporal intent behind user queries, the model supports the integration of temporal information retrieval techniques that prioritize temporal relevance. This focus ensures a better match between relevant web content and users' temporal needs, ultimately enriching the overall search experience. 
Furthermore, when the timestamp of web content is unknown, BiTimeBERT 2.0 can be employed to date this content, aiding in the analysis of transient and rapidly changing web materials, such as distinguishing between old and new content. Additionally, BiTimeBERT 2.0's ability to detect semantic changes over time offers invaluable insights into how concepts develop. 

In summary, the applications of BiTimeBERT 2.0 within the temporal web domain hold great promise for enhancing our understanding of the historical context of web content and streamline temporal web analysis. The model effectively integrates three critical temporal dimensions: document timestamps, content temporal information, and dynamic "Person" entities. This methodological advancement not only provides valuable insights but also fosters innovative approaches to aggregating and analyzing temporal web data, contributing to a deeper understanding of the temporal dimensions inherent in online content.

\section{Conclusions} 

In this paper, we present BiTimeBERT 2.0, a novel time-aware language model that leverages three innovative pre-training objectives: Extended Time-Aware Masked Language Modeling (ETAMLM), Document Dating (DD), and Time-Sensitive Entity Replacement (TSER). Each objective addresses a distinct aspect of temporal information: ETAMLM enhances the model's ability to comprehend temporal contexts and relationships, DD incorporates document timestamps as explicit chronological markers, and TSER captures the temporal dynamics of "Person" entities. Through extensive experiments across diverse time-related tasks, including event occurrence time estimation, document dating, and semantic change detection, we demonstrate that BiTimeBERT 2.0 achieves comparable or superior performance to state-of-the-art models, particularly excelling on datasets that extend beyond the temporal scope of its pre-training corpus. Moreover, our pre-training strategy on the RNYT corpus significantly reduces computational costs, cutting GPU hours by nearly 53\%, making BiTimeBERT 2.0 a more resource-efficient solution. Future research will focus on methods to incorporate newly introduced data dynamically and investigate fine-grained temporal modeling, such as day-level granularity, to further enhance the model's capabilities.

\section*{Acknowledgement}
This research is supported by the Science and Technology Planning Project of Guangdong Province (2020B0101100002), the National Natural Science Foundation of China (62402185, 62476097), the Fundamental Research Funds for the Central Universities, South China University of Technology (x2rjD2240100), the Guangzhou Basic and Applied Basic Research Foundation (Young Doctor "QiHang" Project) (SL2023A04J00753), Guangdong Provincial Natural Science Foundation for Outstanding Youth Team Project (2024B1515040010).  

\FloatBarrier
\bibliographystyle{ACM-Reference-Format}
\bibliography{TimeBERT}
\FloatBarrier

\appendix

\end{document}